\documentclass{article}

\PassOptionsToPackage{numbers, compress, sort}{natbib}

\usepackage[final]{neurips_2021}

\usepackage[utf8]{inputenc} 
\usepackage[T1]{fontenc}    
\usepackage[pagebackref=true, colorlinks=true, allcolors=red, citecolor=green]{hyperref}   
\usepackage{url}            
\usepackage{booktabs}       
\usepackage{amsfonts}       
\usepackage{nicefrac}       
\usepackage{microtype}      
\usepackage{xcolor}         
\usepackage{draftfigure} 

\usepackage{pifont}
\newcommand{\cmark}{\color{green}{\ding{51}}}%
\newcommand{\xmark}{\color{red}{\ding{55}}}%

\usepackage[linesnumbered,ruled,vlined]{algorithm2e}
\usepackage{enumitem}
\usepackage{graphicx}
\usepackage{wrapfig}

\usepackage{amssymb}
\usepackage{amsmath,amsfonts}
\usepackage{amsopn}
\usepackage{bm} 
\usepackage{multirow}
\usepackage{flushend}
\usepackage{tabularx}

\newcommand{\IDOL}{\method{IDOL}\xspace}

\newcommand{\vct}[1]{\boldsymbol{#1}} 

\newcommand{\field}[1]{\mathbb{#1}}
\newcommand{\R}{\field{R}} 

\newcommand{\ProbOpr}[1]{\mathbb{#1}}

\newcommand{\expect}[2]{%
\ifthenelse{\equal{#2}{}}{\ProbOpr{E}_{#1}}
{\ifthenelse{\equal{#1}{}}{\ProbOpr{E}\left[#2\right]}{\ProbOpr{E}_{#1}\left[#2\right]}}} 

\DeclareMathOperator{\argmin}{arg\,min}

\newcommand{\vtheta}{\vct{\theta}}
\newcommand{\vTheta}{\vct{\Theta}}

\newcommand{\vq}{\vct{q}}
\newcommand{\vx}{{\vct{x}}}

\newcommand{\vz}{{\vct{z}}}

\newcommand{\vphi}{\vct{\phi}}

\newcommand{\sS}{\mathcal{S}}

\newcommand{\sT}{\mathcal{T}}

\newcommand{\sU}{\mathcal{U}}
\newcommand{\sL}{\mathcal{L}}
\newcommand{\eat}[1]{}
\newcommand{\method}[1]{\textsc{#1}}

\SetKwInput{KwInput}{Input}                
\SetKwInput{KwOutput}{Output}              
\SetKwInput{KwInit}{Initialize}              

\usepackage{colortbl}
\definecolor{Gray}{gray}{0.9}
\definecolor{LightCyan}{rgb}{0.88,1,1}
\renewcommand{\paragraph}[1]{\vspace{-0.5ex}\textbf{#1}}
\newcommand{\ie}{i.e.\xspace}
\newcommand{\eg}{e.g.\xspace}

\title{Gradual Domain Adaptation \\without Indexed Intermediate Domains}

\author{Hong-You Chen\\
The Ohio State University, USA\\
\texttt{chen.9301@osu.edu}
\And
Wei-Lun Chao\\
The Ohio State University, USA\\
\texttt{chao.209@osu.edu}
}

\begin{document}

\maketitle


\begin{abstract}
The effectiveness of unsupervised domain adaptation degrades when there is a large discrepancy between the source and target domains. Gradual domain adaptation (GDA) is one promising way to mitigate such an issue, by leveraging additional unlabeled data that gradually shift from the source to the target. Through sequentially adapting the model along the ``indexed'' intermediate domains, GDA substantially improves the overall adaptation performance. In practice, however, the extra unlabeled data may not be separated into intermediate domains and indexed properly, limiting the applicability of GDA. In this paper, we investigate how to discover the sequence of intermediate domains when it is not already available. Concretely, we propose a coarse-to-fine framework, which starts with a \emph{coarse} domain discovery step via progressive domain discriminator training. This coarse domain sequence then undergoes a \emph{fine} indexing step via a novel cycle-consistency loss, which encourages the next intermediate domain to preserve sufficient discriminative knowledge of the current intermediate domain. The resulting domain sequence can then be used by a GDA algorithm. On benchmark data sets of GDA, we show that our approach, which we name \textbf{I}ntermediate \textbf{DO}main \textbf{L}abeler (\textbf{\IDOL}), can lead to comparable or even better adaptation performance compared to the pre-defined domain sequence, making GDA more applicable and robust to the quality of domain sequences. Codes are available at \url{https://github.com/hongyouc/IDOL}.
\end{abstract}


\section{Introduction}
\label{s_intro}

The distributions of real-world data change dynamically due to many factors like time, locations, environments, etc. Such a fact poses a great challenge to machine-learned models, which implicitly assume that the test data distribution is covered by the training data distribution. To resolve this generalization problem, unsupervised domain adaptation (UDA), which aims to adapt a learned model to the test domain given its unlabeled data~\cite{gong2012geodesic,ganin2016domain}, has been an active sub-field in machine learning.

Typically, UDA assumes that the ``source'' domain, in which the model is trained, and the ``target'' domain, in which the model is deployed, are discrepant but sufficiently related. Concretely, \citet{zhao2019learning,ben2010theory} show that the generalization error of UDA is bounded by the discrepancy of the marginal or conditional distributions between domains. Namely, the effectiveness of UDA may degrade along with the increase in domain discrepancy. Take one popular algorithm, self-training~\cite{lee2013pseudo,mcclosky2006effective,mcclosky2006reranking}, for example. Self-training adapts the source model by progressively labeling the unlabeled target data (\ie, pseudo-labels) and using them to fine-tune the model~\cite{chen2011co,kim2019self,zou2018unsupervised,khodabandeh2019robust,liang2019distant,inoue2018cross,long2013transfer,zou2019confidence}.
Self-training works if the pseudo-labels are accurate (as it essentially becomes supervised learning), but is vulnerable if they are not, which occurs when there exists a large domain gap~\cite{kumar2020understanding}.

\begin{figure}[t]
    \centering
    {\includegraphics[width=1.10\textwidth]{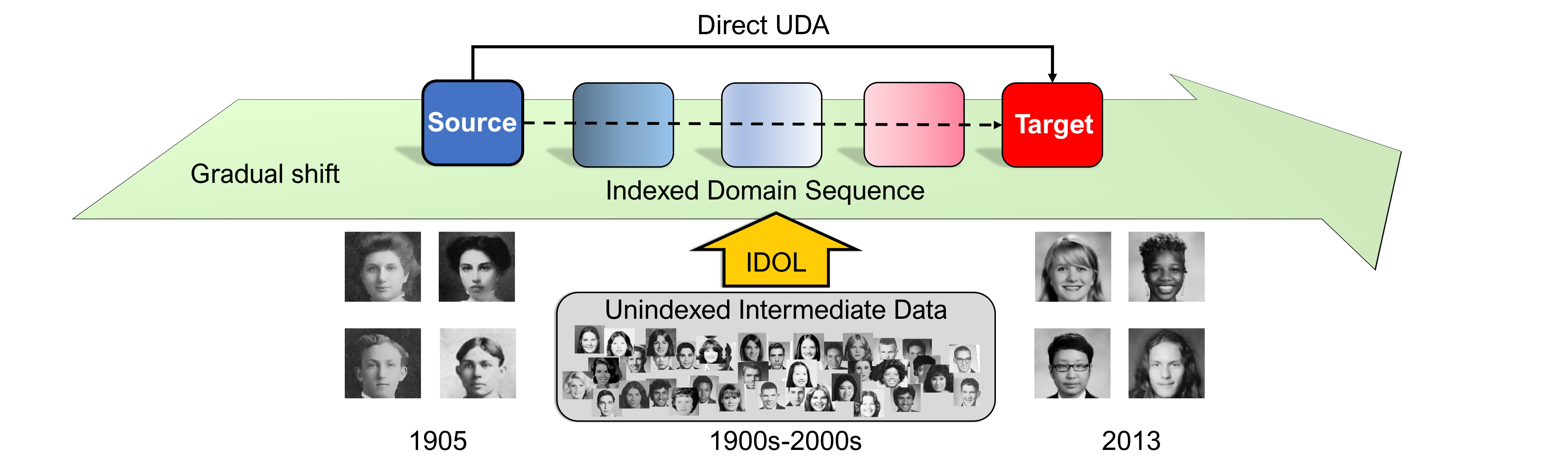}}\hfill
    \vspace{-10pt}
    \caption{\small \textbf{Gradual domain adaptation (GDA) without indexed intermediate domains.} In this setting, one is provided with labeled source, unlabeled target, and additional unlabeled intermediate data that have not been grouped and indexed into a domain sequence. Our approach \textbf{intermediate domain labeler (\IDOL)} can successfully discover the domain sequence, which can then be leveraged by a GDA algorithm to achieve a higher target accuracy than direct unsupervised domain adaptation (UDA). Images are from the Portraits dataset~\cite{ginosar2015century}.}
    \label{fig:overview}
    \vspace{-20pt}
\end{figure}

To address this issue, several recent works investigate \emph{gradual domain adaptation (GDA)}~\cite{hoffman2014continuous,gadermayr2018gradual,wulfmeier2018incremental,kumar2020understanding}, in which beyond the source and target domains, the model can access additional unlabeled data from the intermediate domains that shift gradually from the source to the target. By adapting the model along the sequence of intermediate domains --- \ie, from the ones close to the source to the ones close to the target --- the large domain gap between source and target is chipped away by multiple sub-adaptation problems (between consecutive intermediate domains) whose domain gaps are smaller. Namely, every time the model moves a step closer to the target, instead of taking a huge jump that can significantly decrease the performance (see~\autoref{fig:overview} for an illustration). 
GDA makes sense, as real-world data change gradually more often than abruptly~\cite{vergara2012chemical,farshchian2018adversarial,sethi2017reliable}. The recent work by~\citet{kumar2020understanding} further demonstrates the strength of GDA both empirically and theoretically.

One potential drawback of GDA is the need for a well-defined sequence of intermediate domains. That is, prior to adaptation, the additional unlabeled data must be grouped into multiple domains and indexed with intermediate domain labels that reflect the underlying data distribution shift from the source to the target. Such information, however, may not be available directly. Existing methods usually leverage side information like time tags~\cite{kumar2020understanding,hoffman2014continuous,wulfmeier2018incremental} to define the sequence, which may be sub-optimal. In some applications, even the side information may not be accessible (\eg, due to privacy concerns), greatly limiting the applicability of GDA.

In this paper, we therefore study GDA in the extreme case --- \emph{the additional unlabeled data are neither grouped nor indexed}. Specifically, we investigate how to discover the  ``domain sequence'' from data, such that it can be used to drive GDA. We propose a two-stage coarse-to-fine framework named \textbf{I}ntermediate \textbf{DO}main \textbf{L}abeler (\textbf{\IDOL}). In the first stage, \IDOL labels each intermediate data instance with a \emph{coarse} score that reflects how close it is to the source or target. We study several methods that estimate the distance from a data instance to a domain. We find that a progressively trained domain discriminator --- which starts with source vs. target data but gradually adds data close to either of them into training --- performs the best, as it better captures the underlying data manifold. 

In the second stage, \IDOL then builds upon the \emph{coarse} scores to group data into domains by further considering the discriminative (\eg, classification) knowledge the model aims to preserve from the source to the target. 
Since the additional unlabeled data are fully unlabeled, we take a greedy approach to identify the next intermediate domain (\ie, a group of data) that can best preserve the discriminative knowledge in the current intermediate domain. 
Concretely, we employ self-training~\cite{lee2013pseudo} along the domain sequence discovered so far to provide pseudo-labels for the current domain, and propose a novel \emph{cycle-consistency loss} to discover the next domain, such that the model adapted to the next domain can be ``adapted back'' to the current domain and predict the same pseudo-labels.
The output of \IDOL is a sequence of intermediate domains that can be used by any GDA algorithms~\cite{kumar2020understanding,hoffman2014continuous,wulfmeier2018incremental}.

We validated \IDOL on two data sets studied in~\cite{kumar2020understanding}, including Rotated MNIST~\cite{larochelle2007empirical} and Portraits over years~\cite{ginosar2015century}. \IDOL can successfully discover the domain sequence that leads to comparable GDA performance to using the pre-defined sequence (\ie, by side information). More importantly, \IDOL is compatible with pre-defined sequences --- by treating them as the coarse sequences --- to further improve upon them. We also investigate \IDOL in scenarios where the additional unlabeled data may contain outliers and demonstrate \IDOL's effectiveness even in such challenging cases. To our knowledge, our work is the first to tackle GDA without grouped and indexed intermediate domains. The success of \IDOL opens up broader application scenarios that GDA can contribute to.


\section{Background: gradual domain adaptation (GDA)}
\label{s_background}

\subsection{Setup of UDA and GDA}
We consider an unsupervised domain adaptation (UDA) problem, in which a classifier is provided with labeled data  from the source domain $\sS=\{(\vx^{\sS}_i, y^{\sS}_i)\}_{i=1}^{|\sS|}$ and unlabeled data  from the target domain $\sT = \{\vx^{\sT}_i\}_{i=1}^{|\sT|}$. 
Both $\sS$ and $\sT$ are sampled IID from the underlying joint distributions of the source $P_\sS$ and target $P_\sT$, respectively. 
A source model $\vtheta_\sS$ is typically produced by training on $\sS$ directly. The goal of UDA is to learn a target model $\vtheta_\sT$ such that it can perform well on the target, measured by a loss $\sL(\vtheta_\sT, P_\sT)$. We focus on the standard UDA setup that assumes no label shifts~\cite{combes2020domain}.

\paragraph{Gradual domain adaptation (GDA)}, on top of UDA, assumes the existence of a sequence of domains $P_0, P_1, \ldots P_{M-1}, P_M$, where $P_0$ and $P_M$ are the source domain and target domain, respectively, along with $M-1$ intermediate domains $P_1, \ldots, P_{M-1}$.
The data distributions of domains are assumed to change gradually from the source to the target along the sequence.
One can access unlabeled data $\sU_m = \{\vx^{\sU_m}_i\}_{i=1}^{{|\sU_m|}}$ from each of the intermediate domain, forming a sequence of unlabeled data $\sU_1, \ldots, \sU_{M-1}$. We denote the union of them by $\sU = \{\vx^{\sU}_ i\}_{i=1}^{{|\sU|}}$. For brevity, we assume that each $\sU_m$ is disjoint from the others but the size $|\sU_m|$ is the same $\forall m \in \{1,\cdots,M-1\}$.

\subsection{GDA with pre-defined domain sequences using self-training~\cite{kumar2020understanding}}
\label{ss_GDA_theory}
We summarize the theoretical analysis of GDA using self-training by~\citet{kumar2020understanding} as follows.

\paragraph{Self-training for UDA.}
Self-training~\cite{lee2013pseudo} takes a pre-trained (source) model $\vtheta$ and the target unlabeled data $\sT$ as input.
Denote by $\vz = f(\vx; \vtheta)$ the predicted logits $\vz$, self-training first applies $f(\cdot; \vtheta)$ to target data points $\vx\in\sT$, sharpens the predictions by \texttt{argmax} into pseudo-labels, and then updates the current model $\vtheta$ by minimizing the loss $\ell$ (\eg, cross entropy) w.r.t the pseudo-labels,
\begin{align}
\label{eq:self-training}
& \vtheta_\sT = \texttt{ST}(\vtheta, \sT) = \arg \min_{\vtheta'\in\vTheta} \frac{1}{|\sT|} \sum_{i=1}^{|\sT|} {\ell\big(f(\vx_i; \vtheta'), \texttt{sharpen}(f(\vx_i; \vtheta))\big)},
\end{align}
where $\vtheta_\sT$ denotes the resulting target model and $\texttt{ST}(\vtheta, \sT)$ denotes the self-training process. A baseline use of self-training for UDA is to set $\vtheta$ as the source model $\vtheta_\sS$. However, when the domain gap between $\sS$ and $\sT$ is large, $\vtheta_\sS$ cannot produce accurate pseudo-labels for effective self-training. 

\paragraph{Gradual self-training for GDA.} In the GDA setup, self-training is performed along the domain sequence, from the current model $\vtheta_m$ (updated using $\sU_m$ already) to the next one $\vtheta_{m+1}$ using $\sU_{m+1}$
\begin{align}
& \vtheta_{m+1} = \texttt{ST}(\vtheta_{m}, \sU_{m+1}),
\end{align}
starting from the source model $\vtheta_0=\vtheta_\sS$. By doing so, the model gradually adapts from $\sS$ to $\sT$ through the sequence $\sU_1, \ldots, \sU_{M}$ to produce the final target model $\vtheta_\sT$.
\begin{align}
& \vtheta_{\sT} = \vtheta_{M} = \texttt{ST}\big(\vtheta_{\sS}, (\sU_1, \ldots, \sU_M)\big). \label{eq:gda}
\end{align}
As the domain gap between consecutive domains is smaller than between $\sS$ and $\sT$, the pseudo-labels will be more accurate in each step. As a result, self-training in GDA can lead to a lower target error than in UDA. We note that the domain sequences in~\cite{kumar2020understanding} are pre-defined using side information.

\paragraph{Theoretical analysis.} \citet{kumar2020understanding} make three assumptions for theoretical analysis:
\begin{itemize}[nosep,topsep=0pt,parsep=0pt,partopsep=0pt, leftmargin=*]
    \item Separation: for every domain $P_m$, the data are separable such that there exists an $R$-bounded (linear) classifier $\vtheta_m \in \vTheta$ with $\|\vtheta_m\|_2\leq R$ that achieves a low loss of $\sL(\vtheta_m, P_m)$.
    \item Gradual shift: for the domain sequence $P_0,\ldots,P_m$ with no label shifts, the maximum per-class Wasserstein-infinity distance $\rho(P_m, P_{m+1})$ between consecutive domains should be $\leq\rho<\frac{1}{R}$.
    \item Bounded data: finite samples $X$ from each domain $P_m$ are bounded, \ie, $E_{X\sim P_m}[\|X\|^2_2]\leq B^2$.
\end{itemize}
Under these assumptions of GDA, if the source model $\theta_0$ has a low loss $\sL(\theta_0, P_0)$ on $P_0$, then with a probability $\delta$, the resulting target model $\theta_M$ through gradual self-training has
\begin{align} \label{eq:error_bound}
\sL(\vtheta_M, P_M) \leq \beta^{M+1} \Big( \sL(\vtheta_0, P_0) + \frac{4BR + \sqrt{2 \log{2M / \delta}}}{\sqrt{|\sU|}} \Big), \hspace{10pt} \text{where } \beta = \frac{2}{1-\rho R}.
\end{align}
That is, the target error is  controlled by the intermediate domain shift $\rho$. If we can sample infinite data (\ie, $|\sU|\rightarrow \infty$) and the source classifier is perfect (\ie, $\sL(\theta_0, P_0)=0$), with a small distance $\rho(P_m, P_{m+1})\leq\rho$ between consecutive domains, gradual self-training achieves a zero target error. 


\section{GDA without indexed intermediate domains}
\label{s_approach}

\subsection{Setup and motivation}
In this paper, we study the case of GDA where the additional unlabeled data $\sU$ are not readily divided into domain sequences $\sU_1, \cdots, \sU_{M-1}$. In other words, we are provided with a single, gigantic unlabled set that has a wide range of support from the source to the target domain. One naive way to leverage it is to adapt from $\sS$ to $\sU$\footnote{In~\cite{kumar2020understanding}, the authors investigated sampling $\sU_1, \cdots, \sU_{M-1}$ from $\sU$ uniformly at random with replacement.} 
and then to $\sT$, which however leads to a much larger target error than applying self-training along the properly indexed intermediate domains~\cite{kumar2020understanding}. We attribute this to the large $\rho(\sS, \sU)$ or $\rho(\sU, \sT)$ according to~\autoref{eq:error_bound}. To take advantage of $\sU$, we must separate it into intermediate domains such that $\rho(P_m, P_{m+1})$ is small enough for every $m\in\{0,\ldots,M-1\}$.

\subsection{Overview of our approach: intermediate domain labeler (\IDOL)}

Given the source data $\sS$, the target data $\sU_M = \sT$, and the unlabeled intermediate data $\sU= \{\vx^{\sU}_ i\}_{i=1}^{{|\sU|}}$, we propose the intermediate domain labeler (\IDOL), whose goal is to sort intermediate data instances in $\sU$ and chunk them into $M-1$ domains $\sU_1, \cdots, \sU_{M-1}$ (with equal sizes) for GDA to succeed.

Taking gradual self-training in \autoref{eq:gda} as an example, \IDOL aims to solve the following problem
\begin{align}
& \min \sL(\vtheta_{\sT}, P_{\sT}),\nonumber\\
& \text{s.t. }\vtheta_{\sT} = \texttt{ST}\big(\vtheta_{\sS}, (\sU_1, \ldots, \sU_M)\big) \quad\text{and}\quad  (\sU_1, \ldots, \sU_{M-1}) = \IDOL(\sS, \sT, \sU; M-1),
\label{eq:IDOL}
\end{align}
where the target model $\vtheta_{\sT}$ is produced by applying gradual self-training to the domain sequence discovered by \IDOL. \IDOL accepts the number of intermediate domains $M-1$ as a hyper-parameter.

Solving \autoref{eq:IDOL} is hard, as evaluating any output sequence needs running through the entire gradual self-training process, not to mention that we have no labeled target data to estimate $\sL(\vtheta_{\sT}, P_{\sT})$. In this paper, we propose to solve \autoref{eq:IDOL} approximately via a coarse-to-fine procedure.

\paragraph{The coarse stage.} We aim to give each $\vx^{\sU}_i\in\sU$ a score $q_i$, such that it tells $\vx$'s position in between the source and target. Specifically, a higher/lower $q_i$ indicates that $\vx^{\sU}_i$ is closer to the source/target domain. With these scores, we can already obtain a coarse domain sequence by \emph{sorting them in the descending order and dividing them into $M-1$ chunks}. 

\paragraph{The fine stage.} Instead of creating the domain sequence $\sU_1, \cdots, \sU_{M-1}$ by looking at the score of each individual $\vx^{\sU}_i\in\sU$, we further consider how data grouped into the same domain can collectively preserve the discriminative knowledge from the previous domains --- after all, the goal of UDA or GDA is to pass the discriminative knowledge from the labeled source domain to the target domain. To this end, we propose a novel cycle-consistency loss to refine the coarse scores progressively.

\subsection{The coarse stage: assigning domain scores}\label{ss_index_func}

In this stage, we assign each $\vx^{\sU}_i\in\sU$ a score $q_i$, such that higher/lower $q_i$ indicates that $\vx^{\sU}_i$ is closer to the source/target domain. We discuss some design choices. For brevity, we omit the superscript $^{\sU}$.

\paragraph{Confidence scores by the classifier $f(\vx; \vtheta)$.} We first investigate scoring each intermediate data instance by the confidence score of the source model $\vtheta_{\sS}$, \ie, $q_i = \max f(\vx_i; \vtheta_{\sS})$ across classes. This is inspired by outlier detection~\cite{liang2017enhancing} and the common practice of self-training, which selects only data with sufficiently large confidence scores to fine-tune upon~\cite{lee2013pseudo,kumar2020understanding}. Here we employ an advanced version, which is to use the latest $\vtheta$ (\ie, the model adapted to the current intermediate domain) to select the next intermediate domain. That is, every time we select the highest confidence instances from the \emph{remaining} ones in $\sU$ to adapt $\vtheta$. Overall, we can give data selected in the $m$-th round ($m\in\{1,\cdots,M-1\}$) a score $\frac{M-1-m}{M-2}\in [0, 1]$.
One drawback of this method is the blindness to the target domain $\sT$. Specifically, we find that using confidence scores tends to select easily classified examples, which do not necessarily follow the gradual shift from the source to the target.

\paragraph{Manifold distance with the source features.}
To model the flow from the source to the target, we argue that it is crucial to consider both sides as references. We thus extract features of $\sS$, $\sT$, and $\sU$ using the source model $\vtheta_\sS$ and apply a manifold learning algorithm (\eg, UMAP~\cite{mcinnes2018umap}) to discover the underlying manifold.
Denote by $\gamma(\vx)$ the dimension-reduced features of $\vx$ after UMAP, we compute the ratio of its distance to the nearest point in the $
\sT$ and $\sS$ as the score $q_i = \frac{\min_{\vx^\sT\in\sT}\|\gamma(\vx_i) - \gamma(\vx^\sT)\|_2}{\min_{\vx^\sS\in\sS}\|\gamma(\vx_i) - \gamma(\vx^\sS)\|_2}$.   

\paragraph{Domain discriminator.}
We investigate another idea to emphasize the roles of the source and target, which is to train a domain discriminator~\cite{ganin2016domain}.
Concretely, we construct a binary classifier $g(\cdot; \vphi)$ using deep neural networks, which is trained to separate the source data $\sS$ (class: 1) and the target data $\sT$ (class: 0). We use a binary-cross entropy loss to optimize the learnable parameter $\vphi$
\begin{align}
\sL(\vphi) = -\frac{1}{|\sS|}\sum_{\vx^\sS\in\sS}\log(\sigma(g(\vx^\sS; \vphi))) - \frac{1}{|\sT|}\sum_{\vx^\sT\in\sT} \log(1-\sigma(g(\vx^\sT; \vphi))), \label{e_d_disc}
\end{align}
where $\sigma$ is the sigmoid function. We can then assign a score to $\vx_i\in\sU$ by $q_i = g(\vx_i; \vphi)$.

\paragraph{Progressive training for the domain discriminator.}
The domain discriminator $g(\cdot; \vphi)$, compared to the manifold distance, better contrasts the source and target domains but does not leverage any of the data $\vx\in\sU$. That is, it might give a faithful score to an example $\vx$ that is very close to $\sS$ or $\sT$, but for other examples that are far away and hence out-of-domain from both ends, the score by $g(\cdot; \vphi)$ becomes less reliable, not to mention that neural networks are usually poorly calibrated for these examples~\cite{guo2017calibration,liang2017enhancing}. We therefore propose to progressively augment the source and target data with $\vx\in\sU$ that has either a fairly high or low $g(\vx; \vphi)$, and fine-tune the domain discriminator $g(\vx; \vphi)$. We perform this process for $K$ rounds, and every time include $\frac{|\sU|}{2K}$ new examples into the source and target. By doing so, the domain discriminator $g(\vx; \vphi)$ is updated with data from $\sU$ and thus can provide more accurate scores to distinguish the remaining examples into the source or target side. {
More specifically, let $N$ denote $|\sU|$, we repeat the following steps for $k$ from $1$ to $K$:

\begin{enumerate}[nosep,topsep=0pt,parsep=0pt,partopsep=0pt, leftmargin=*]
\item Train $g(\cdot, \vphi)$ using $\sS$ and $\sT$, based on the loss in \autoref{e_d_disc}.
\item Predict $\hat{q}_i = g(\vx_i, \vphi), \forall \vx_i \in \sU$.
\item Rank all $\hat{q}_i$ in the descending order.
\item The data with the $\frac{N}{2K}$ largest $\hat{q}_i$ scores form a new intermediate domain for the source side (their $q_i$ is set to $\frac{2K-k}{2K}$). We remove them from $\sU$ and add them into $\sS$.
\item The data with the $\frac{N}{2K}$ smallest $\hat{q}_i$ scores form a new intermediate domain for the target side (their $q_i$ is set to $\frac{k}{2K}$). We remove them from $\sU$ and add them into $\sT$.
\end{enumerate}
}

Overall, we give a data instance $\vx \in \sU$ selected in the $k$-th round ($k\in\{1,\cdots,K\}$) a score $\frac{2K -k}{2K}$ if it is added into the source side, or $\frac{k}{2K}$ if it is added into the target side.


\subsection{The fine stage: cycle-consistency for refinement}\label{ss_metaGDA}
The domain scores developed in \autoref{ss_index_func} can already give each individual example $\vx\in\sU$ a rough position of which intermediate domain it belongs to. However, for GDA to succeed, we must consider intermediate data in a collective manner. That is, each intermediate domain should preserve sufficient discriminative (\eg, classification) knowledge from the previous one, such that after $M$ rounds of adaptation through GDA, the resulting $\vtheta_{\sT}$ in~\autoref{eq:IDOL} will be able to perform well on $\sT$.  
This is reminiscent of some recent claims in UDA~\cite{zhao2019learning,long2017conditional}: matching only the marginal distributions  between domains (\ie, blind to the class labels) may lead to sub-optimal adaptation results.

\emph{But how can we measure the amount of discriminative knowledge preserved by an intermediate domain (or in the target domain) if we do not have its labeled data?} To seek for the answer, we revisit \autoref{eq:error_bound} and consider the case $M=1$. If the two domains $P_0$ and $P_1$ are sufficiently close and the initial model $\vtheta_0$ is well-trained, the resulting model $\vtheta_1$ after self-training will perform well. Reversely, we can treat such a well-performing $\vtheta_1$ as the well-trained initial model, and apply self-training again --- this time from $P_1$ back to $P_0$. The resulting model $\vtheta_0'$, according to the same rationale, should perform well on $P_0$. In other words, $\vtheta_0'$ and $\vtheta_0$ should predict similarly on data sampled from $P_0$. The \textbf{similarity between $\vtheta_0'$ and $\vtheta_0$ in terms of their predictions} therefore can be used as a proxy to measure \textbf{how much discriminative knowledge $\vtheta_1$, after adapted to $P_1$, preserves}. It is worth noting that measuring the similarity between $\vtheta_0'$ and $\vtheta_0$ requires no labeled data from $P_0$ or $P_1$.

Now let us consider a more general case that $M$ is not limited to $1$. Let us set $\vtheta_0 = \vtheta_{\sS}$, where $\vtheta_{\sS}$ has been trained with the labeled source data $\sS$ till convergence. Based on the concept mentioned above, we propose a novel learning framework to discover the domain sequences,
\begin{align}
\argmin_{(\sU_1, \ldots, \sU_{M-1})} & \expect{\vx\sim P_{0}}{ \ell\big(f(\vx; \vtheta_0'), \texttt{sharpen}(f(\vx; \vtheta_0))\big)},  \nonumber\\
\vspace{10pt}\text{s.t., } & \vtheta'_{0} = \texttt{ST}(\vtheta_{M}, (\sU_{M-1}, \ldots, \sU_0)), \label{eq:reverse_gda}\\
& \vtheta_{M} = \texttt{ST}(\vtheta_{0}, (\sU_1, \ldots, \sU_M)), \nonumber
\end{align} 
where $\sU_0$ is the source data $\sS$ with labels removed, and $\sU_M=\sT$ is the target data. $\vtheta_0'$ is a gradually self-trained model from $\vtheta_0$ that undergoes a cycle   $\sU_1 \rightarrow, \cdots, \rightarrow \sU_M \rightarrow,  \cdots, \rightarrow \sU_0$.
In other words, \autoref{eq:reverse_gda} aims to find the intermediate domain sequences such that performing gradual self-training along it in a forward and then backward fashion leads to cycle-consistency~\cite{zhu2017unpaired}.

\begin{figure}[t]
    \centering
    \hspace{-5pt}
    {\includegraphics[width=0.595\textwidth]{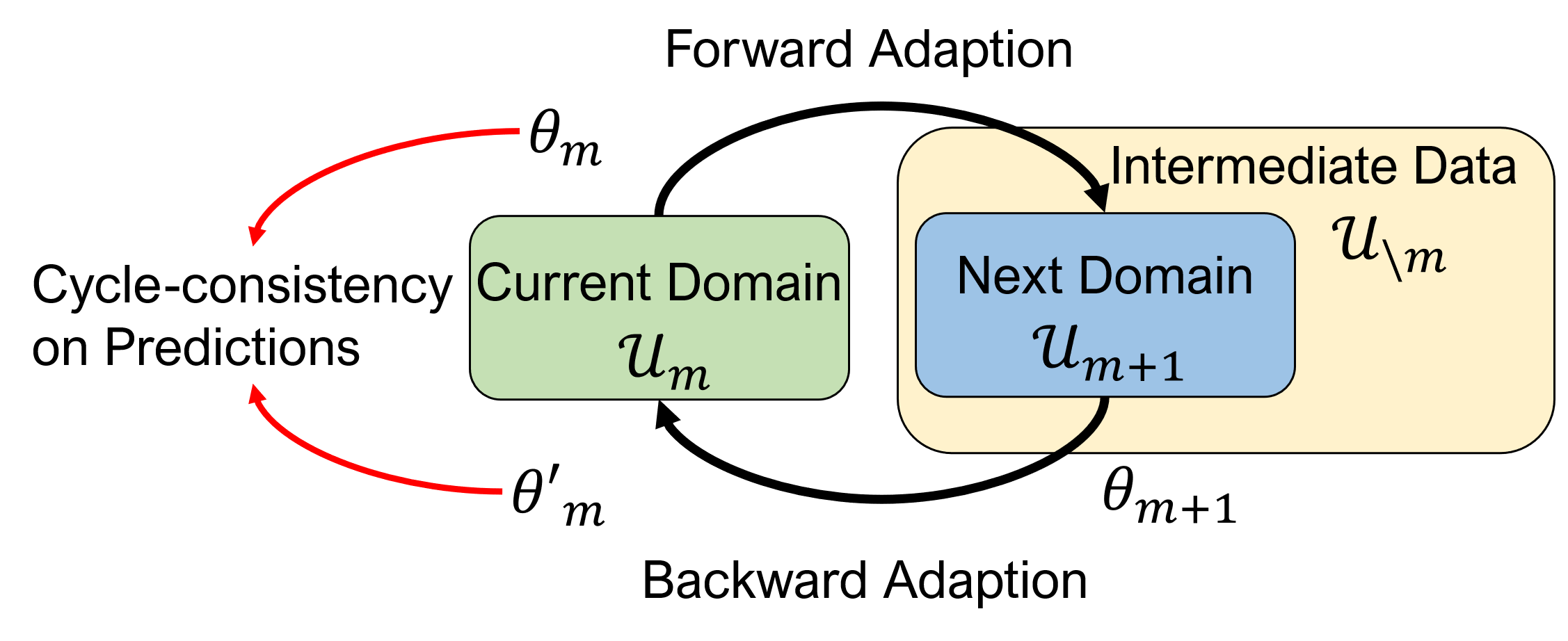}}%
    \vskip -10pt
    \caption{\small Cycle-consistency (cf. \autoref{eq:reverse_gda_one}).}
    \label{fig:reverse_GDA}    
    \vskip -10pt
\end{figure}

\textbf{Optimization.} Solving~\autoref{eq:reverse_gda} is still not trivial due to its combinatorial nature. We therefore propose to solve it \textbf{greedily, starting with discovering $\sU_1$, and then $\sU_2$, and then so on.} Concretely, we decompose \autoref{eq:reverse_gda} into a series of sub-problems, as illustrated in~\autoref{fig:reverse_GDA}, each is defined as
\begin{align}
\argmin_{\sU_{m+1}\subset\sU_{\setminus m}} \quad &  \frac{1}{|\sU_{m}|}\sum_{\vx\in\sU_{m}} { \ell\big(f(\vx; \vtheta_m'), \texttt{sharpen}(f(\vx; \vtheta_m))\big)},  \nonumber\\
\vspace{10pt}\text{s.t., } \quad & \vtheta'_{m} = \texttt{ST}(\vtheta_{m+1}, \sU_{m}), \label{eq:reverse_gda_one}\\
& \vtheta_{m+1} = \texttt{ST}(\vtheta_{m}, \sU_{m+1}), \nonumber
\end{align}
where $\vtheta_m$ is the model already adapted to the current domain $\sU_m$, and $\sU_{\setminus m} = \sU\setminus \cup_{j=1}^m \sU_j$ is the remaining data that have not been selected by the $m$ intermediate domains so far.
That is, each sub-problem {discovers only the next intermediate domain $\sU_{m+1}$} via cycle-consistency.

Let $N=|\sU_{\setminus m}|$ and $\sU_{\setminus m} = \{\vx_i\}_{i=1}^N$, selecting $\sU_{m+1}\subset\sU_{\setminus m}$ is equivalent to selecting a binary indicator vector $\vq\in\{0, 1\}^N$, in which $q_i=1$\footnote{Here we use the same notation as the coarse scores defined in \autoref{ss_index_func}, since later we will initialize these values indeed by the coarse scores.} means that  $\vx_i\in\sU_{\setminus m}$ is included into $\sU_{m+1}$. This representation turns the self-training step $\vtheta_{m+1} = \texttt{ST}(\vtheta_{m}, \sU_{m+1})$ in~\autoref{eq:reverse_gda_one} into
\begin{align}
& \texttt{ST}(\vtheta_{m}, \vq) = \arg \min_{\vtheta\in\vTheta} \frac{1}{N} \sum_{i=1}^{N} q_i\times{\ell\big(f(\vx_i; \vtheta), \texttt{sharpen}(f(\vx_i; \vtheta_m))\big)},    
\end{align}
We choose to solve \autoref{eq:reverse_gda_one} approximately by relaxing the binary vector $\vq\in\{0, 1\}^N$ into a real vector $\vq\in\R^N$, which fits perfectly into the meta-reweighting framework~\cite{ren2018learning,jamal2020rethinking}. Concretely, meta-reweighting treats $\vq\in\R^N$ as learnable differentiable parameters associated to training examples (in our case, the data in $\sU_{\setminus m}$). Meta-reweighting for \autoref{eq:reverse_gda_one} can be implemented via the following six steps for multiple iterations. 
\begin{enumerate}[nosep,topsep=0pt,parsep=0pt,partopsep=0pt, leftmargin=*]
\item Detach: $\vtheta\leftarrow \vtheta_m$,
\item Forward: $\vtheta(\vq) \leftarrow \vtheta - \cfrac{\eta_{\vtheta}}{|\sU_{\setminus m}|}\times \cfrac{\partial\sum_{i\in\sU_{\setminus m}} q_i\times\ell(f(\vx_i; \vtheta), \texttt{sharpen}(f(\vx_i; \vtheta_m)))}{\partial \vtheta},$
\item Detach: $\vtheta'\leftarrow \vtheta(\vq)$,
\item Backward: $\vtheta(\vq) \leftarrow \vtheta(\vq) -\cfrac{\eta_{\vtheta}}{|\sU_m|}\times \cfrac{\partial\sum_{j\in \sU_m} \ell(f(\vx_j; \vtheta(\vq)), \texttt{sharpen}(f(\vx_j; \vtheta')))}{\partial \vtheta(\vq)},$
\item Update: $\vq \leftarrow \vq - \cfrac{\eta_{\vq}}{|\sU_m|}\times \cfrac{\partial\sum_{j\in \sU_m} \ell(f(\vx_j; \vtheta(\vq)), \texttt{sharpen}(f(\vx_j; \vtheta_m)))}{\partial \vq}$,
\item Update: $q_i \leftarrow \max\{0, q_i\}$.
\end{enumerate}
The aforementioned updating rule gives $\vx_i$ a higher value of $q_i$ if it helps preserve the discriminative knowledge. After obtaining the updated $\vq\in \R^N$, we then sort it in the descending order and select the top $\frac{|\sU|}{M-1}$ examples to be $\sU_{m+1}$ (\ie, every intermediate domain has an equal size). 

\textbf{Discussion.} {The way we relax the binary-valued vector $\vq\in\{0, 1\}^N$ by real values is related to the linear programming relaxation of an integer programming problem. It has been widely applied, for example, in discovering sub-domains within a dataset~\cite{gong2013reshaping}. Theoretically, we should constrain each element of $\vq$ to be within $[0, 1]$. Empirically, we found that even without the clipping operation to upper-bound $q_i$ (\ie, just performing $\max{\{0, q_i\}}$), the values in $q_i$ do not explode and the algorithm is quite stable: we see a negligible difference of using upper-bounded $q_i$ or not in the resulting accuracy of GDA. This is also consistent with the common practice of using meta-reweighting~\cite{jamal2020rethinking,ren2018learning}.} 

\textbf{Initialization.} We initialize $q_i$ by the coarse scores defined in \autoref{ss_index_func}. This is for one important reason: \autoref{eq:reverse_gda_one} searches for the next intermediate domain $\sU_{m+1}$ only based on the current intermediate domain $\sU_{m}$.
Thus, it is possible that $\sU_{m+1}$ will include some data points that help preserve the knowledge in $\sU_{m}$ but are distributionally deviated from the gradual transition between $\sU_{m}$ and the target domain $\sT$.
The scores defined in \autoref{ss_index_func} thus serve as the guidance; we can also view \autoref{eq:reverse_gda_one} as a refinement step of the coarse scores. As will be seen in \autoref{s_exp}, the qualities of the initialization and the refinement steps both play important roles in \IDOL's success.


\section{Experiment}
\label{s_exp}

\subsection{Setup}
We mainly study two benchmark datasets used in~\cite{kumar2020understanding}. Rotated MNIST~\cite{larochelle2007empirical} is a popular synthetic dataset to study distributional shifts. It gradually rotates the original $50,000$ images in the MNIST~\cite{lecun1998gradient} dataset with $[0, 60]$ degrees uniformly, where $[0, 5)$/$[5, 55)$/$[55, 60]$ degrees are for the source/intermediate/target domains, respectively. However, since original images have no ground truths of the rotation, the rotation annotations are not expected to be perfect. Portraits dataset~\cite{ginosar2015century} is a real-world gender classification dataset of a collection of American high school seniors from the year 1905 to 2013. Naturally, the portrait styles change along years (\eg, lip curvature~\cite{ginosar2015century}, fashion styles). The images are first sorted by the years and split. For both datasets, each domain contains $2000$ images, and $1,000$ images are reserved for both the source and target domains for validation. 

We follow the setup in~\cite{kumar2020understanding}: each model is a convolutional neural network trained for $20$ epochs for each domain consequently (including training on the source data), using Adam optimizer~\cite{kingma2014adam} with a learning rate $0.001$, batch size $32$, and weight decay $0.02$. We use this optimizer as the default if not specified. Hyper-parameters of \IDOL include $K=2M$ rounds for progressive training and $30$ epochs of refinement per step (with mini-batch $128$), where $M=19$ for the Rotated MNIST and $M=7$ for the Portraits. More details are in the supplementary material. 
  
To study how the intermediate domain sequences affect gradual self-training, we consider baselines including ``{Source only}'' and ``{UDA}'' that directly self-trains on the target  data ($\sT$) and all the intermediate data ($\sU$). We compare different domain sequences including \emph{Pre-defined} indexes, \eg, \emph{rotations} and \emph{years}. We further consider the challenging setup that all intermediate data are unindexed. \emph{Random} sequences are by dividing the unlabeled data into $M-1$ domains randomly, with self-training on the target in the end. For our \IDOL, different domain scores introduced in~\autoref{ss_index_func} are adopted for the coarse domain sequences, and we optionally apply cycle-consistency refinement to make them fine-grained.

\begin{figure}[t]
    \centering
    \vspace{-5pt}
    \minipage{1.0\textwidth}
    \minipage{0.08\textwidth}
    {\includegraphics[height=2.4cm,keepaspectratio]{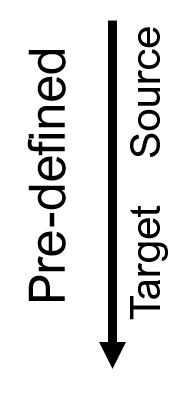}}%
    \endminipage\hfill
    \minipage{0.53\textwidth}
    {\includegraphics[height=2.8cm,keepaspectratio]{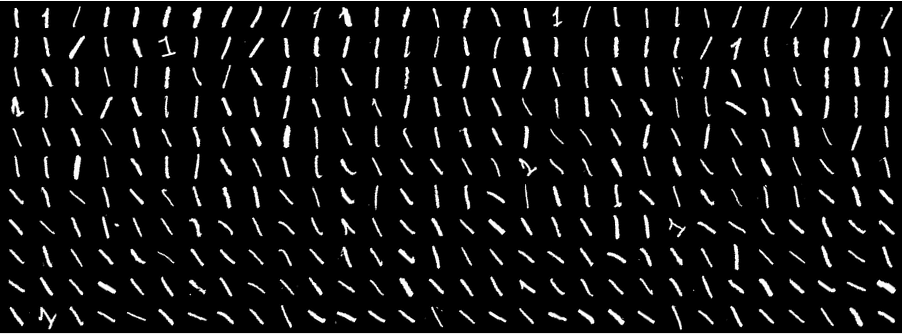}}%
    \centering
    \mbox{}
    \endminipage\hfill
    \minipage{0.35\textwidth}
    {\includegraphics[height=2.8cm,keepaspectratio]{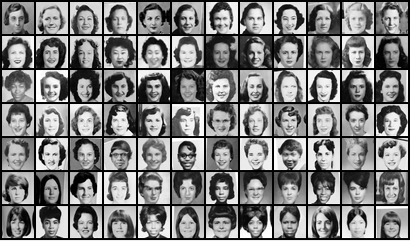}}%
    \centering
    \mbox{}
    \endminipage\hfill
    \endminipage\vfill
    \minipage{1.0\textwidth}
    \minipage{0.08\textwidth}
    {\includegraphics[height=2.4cm,keepaspectratio]{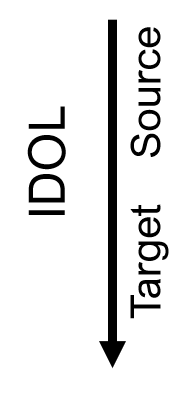}}%
    \endminipage\hfill
    \minipage{0.53\textwidth}
    {\includegraphics[height=2.8cm,keepaspectratio]{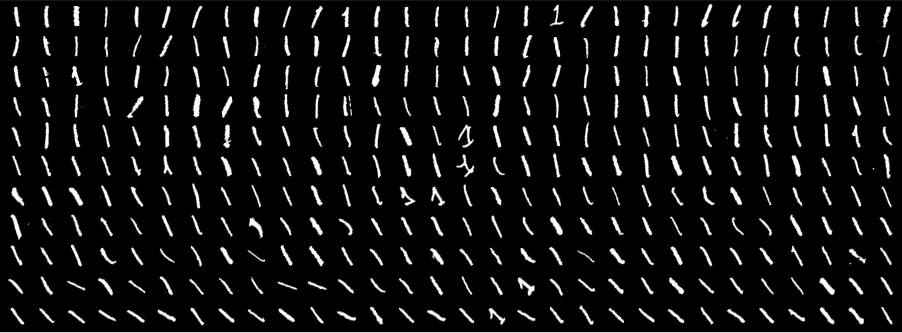}}%
    \centering\vfill
    \mbox{\small (a) Rotated MNIST (class ``1'')}
    \endminipage\hfill
    \minipage{0.35\textwidth}
    {\includegraphics[height=2.8cm,keepaspectratio]{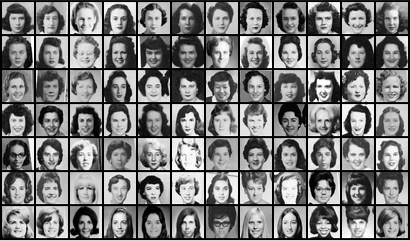}}%
    \centering\vfill
    \mbox{\small (b) Portraits (class ``female'')}
    \endminipage
    \endminipage 
    \caption{\small Random samples (the most representative class) from pre-defined indexes or \IDOL sequences. }  
    \label{fig:example} 
    \vspace{-5pt}
\end{figure}

\begin{table}[t] 
    \footnotesize
    \small
	\centering
	\tabcolsep 4.5pt
	\renewcommand{\arraystretch}{0.7}
	\caption{\small Gradual self-training on Rotated MNIST and Portraits. Bottom section: \IDOL with domain scores.}
	\begin{tabular}{l|cc|c|cc}
	\toprule
	Coarse scores & Indexed? & Adaptation & Refined?  & Rotated MNIST & Portraits\\
	\midrule    
	\multirow{3}{*}{None} & \multirow{3}{*}{\xmark}  & Source only& -  & 31.9$\pm$1.7 & 75.3$\pm$1.6\\
	 &   &UDA ($\sT$) &-  & 33.0$\pm$2.2 & 76.9$\pm$2.1 \\
	 &  & UDA ($\sT+\sU$)  & - & 38.0$\pm$1.6 & 78.9$\pm$3.0 \\
    \midrule
    \multirow{2}{*}{Pre-defined~\cite{kumar2020understanding} } & \multirow{2}{*}{\cmark} & \multirow{2}{*}{GDA} & \xmark  & 87.9$\pm$1.2 & 83.8$\pm$0.8\\
    &  & & \cmark  & 93.3$\pm$2.3 & 85.8$\pm$0.4\\
    \midrule
    \midrule
    \multirow{2}{*}{Random} & \multirow{2}{*}{\xmark} & \multirow{2}{*}{GDA} & \xmark  & 39.5$\pm$2.0 & 81.1$\pm$1.8\\
     &  & &\cmark & 57.5$\pm$2.7 & 82.5$\pm$2.2 \\
    \midrule
    Classifier confidence & \multirow{5}{*}{\xmark} & \multirow{6}{*}{GDA} & \xmark & 45.5$\pm$3.5 & 79.3$\pm$1.7\\
    Manifold distance &  & &\xmark & 72.4$\pm$3.1 & 81.9$\pm$0.8\\
    Domain discriminator &  & &\xmark& 82.1$\pm$2.7 & 82.3$\pm$0.9\\
    \multirow{2}{*}{Progressive domain discriminator}&   & &\xmark  & 85.7$\pm$2.7 & 83.4$\pm$0.8\\
    &   & &\cmark & 87.5$\pm$2.0 & 85.5$\pm$1.0\\
    \bottomrule
	\end{tabular}
	\vspace{0.1cm} 
	\label{tbl:main}
	\vspace{-0.6cm} 
\end{table}

\subsection{Main study: comparison with pre-defined domain sequences}\label{exp:main_study}
The main results on Rotated MNIST and Portraits are provided in~\autoref{tbl:main}. We first verify that without any domain sequences, UDA (on target and intermediate data) and GDA (on random sequences) do not perform well, though they slightly improve over the source model. On the contrary, training with pre-defined indexes that arguably guide the gradual adaptation significantly performs better.

\paragraph{\IDOL is competitive to pre-defined sequences.}
\IDOL can produce meaningful domain sequences (see  \autoref{fig:example}), \emph{given only the intermediate unlabeled data without any order}. The domain discriminator with progressive training performs the best and is comparable to pre-defined sequences. The confidence scores by the classifier are not enough to capture domain shifts.

\begin{wrapfigure}{r}{0.35\textwidth}
    \centering
    \minipage{0.45\textwidth}
    {\centering\includegraphics[width=0.65\textwidth]{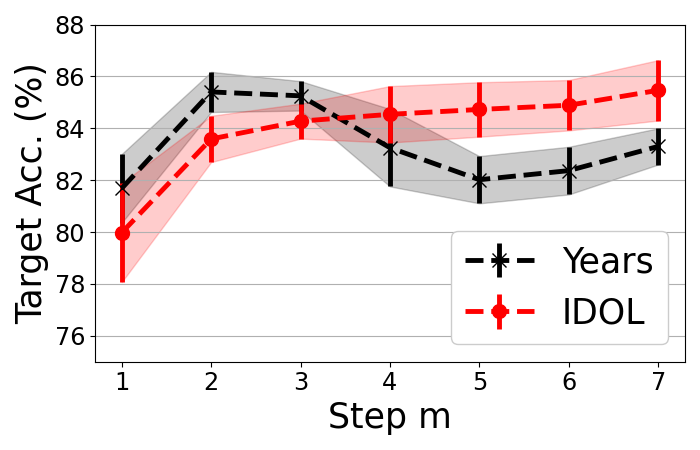}}%
    \endminipage \hfill
    \minipage{0.45\textwidth}
    \mbox{\small (a) Target Acc. of gradual ST.}\hfill
    \hspace{-20pt}
    {\centering\includegraphics[width=0.65\textwidth]{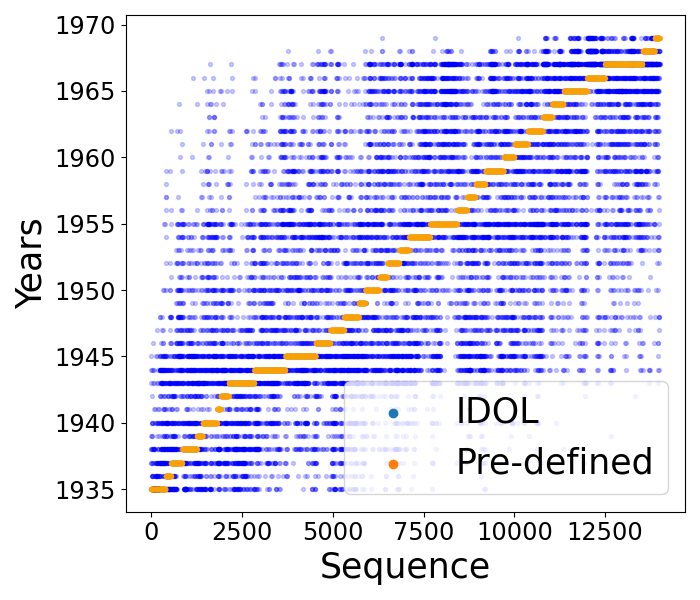}}
    \vspace{-5pt}
    \mbox{\small (b) Years vs. \IDOL sequence.}\hfill
    \endminipage 
    \vspace{5pt}
    \caption{\small Portraits with year indexes or \IDOL domain sequence.}  
    \label{fig:por_step} 
    \vspace{-15pt}
\end{wrapfigure}

\paragraph{Refinement helps and coarse initialization matters.}
We observe that it is crucial to have high-quality coarse sequences as the initialization; the fine indexes by cycle-consistency refinement could further improve the coarse sequences, validating its effectiveness. 
{We note that the Rotated MNIST dataset treats the original MNIST data as 0-degree rotation and artificially rotates the data given a rotation index. However, for data in the original MNIST dataset, there already exist variations in terms of rotations. This can be seen in~\autoref{fig:example}: in the first row of $0$-degree rotation based on the pre-defined indexes, the examples of digit $1$ do have slight rotations. Interestingly, in this situation, \IDOL could potentially capture the true rotation of each example to further improve GDA.}

\paragraph{Analysis.} 
To understand why refinement helps, we take a closer look at the Portraits dataset that is naturally sorted by \emph{years}. Are \emph{years} the best factor to construct the domain sequence? We compare it with the refined sequence by \IDOL. In~\autoref{fig:por_step}, we monitor the target accuracy of each step of adaptation and find that the accuracy fluctuates by using years. Interestingly, learning with \IDOL stably and gradually improves the target accuracy and ends up with a higher accuracy, validating that refinement with cycle-consistency indeed produces better sequences. {Specifically, the \IDOL sequence has a reasonable $0.727$ correlation with \emph{years} but they are not perfectly aligned. Several factors such as fashions, hairstyles, eyeglasses, and lip curvatures may not perfectly align with years~\cite{ginosar2015century}. Even within the same year, individuals could have variations in these factors as well. We thus hypothesize that \IDOL can discover the domain sequence that reflects a smoother transition of these factors.}

We note that the intermediate domains constructed by IDOL (domain scores with or without refinement) may not match the target ``class'' distributions. We monitor the number of examples of class $c$ in each domain $\sU_m$, \ie, $|\sU_{m,c}|$, and compute $\frac{1}{M-1}\sum_{m=1}^{M-1}\frac{\max_c{\{|\sU_{m,c}|\}}}{\min_c{\{|\sU_{m,c}|\}}}$, which ideally should be $1.0$, \ie,  class-balanced. We observe fine-grained indexes are indeed more class-balanced (coarse vs fine-grained): $1.33$ vs $1.23 $ on Rotated MNIST and $1.27$ vs $1.25$ on Portraits.

\subsection{Case study: learning with partial or outlier information}
If partial or outlier information of the intermediate domains is given, is \IDOL still helpful? We examine the question by three experiments. First, we consider \emph{unsorted domains}: the data are grouped into $M-1$ domains according to the pre-defined indexes but the domains are \emph{unordered}. We find that if we sort the domains with the mean domain scores $\frac{1}{|\sU_m|}\sum_{\vx_i \in{\sU_m}} q_i$, then we can perfectly recover the pre-defined order. Second, we investigate if we have \emph{fewer, coarsely separated pre-defined domains}: 
we double the size of $|\sU_m|$. We find that on Rotated MNIST (w/ pre-defined indexes), the accuracy degrades by $11\%$. \IDOL can recover the accuracy since \IDOL outputs a sorted sequence of data points, from which we can construct more, fine-grained intermediate domains. Third, we investigate the addition of \emph{outlier domains}, in which the starting/end intermediate domains do not match the source/target domains exactly. For example, on Rotated MNIST, 
the intermediate domains are expanded to cover $[-30, 90]$ degrees. GDA on such domain sequence with outlier domains degrades to $77.0\pm2.0$ accuracy. Refinement can still improve it to $81.3\pm1.4$.

\begin{wrapfigure}{r}{0.3\textwidth}
    \centering
    \vspace{-15pt}
    \minipage{0.5\textwidth}
    {\includegraphics[width=0.6\textwidth]{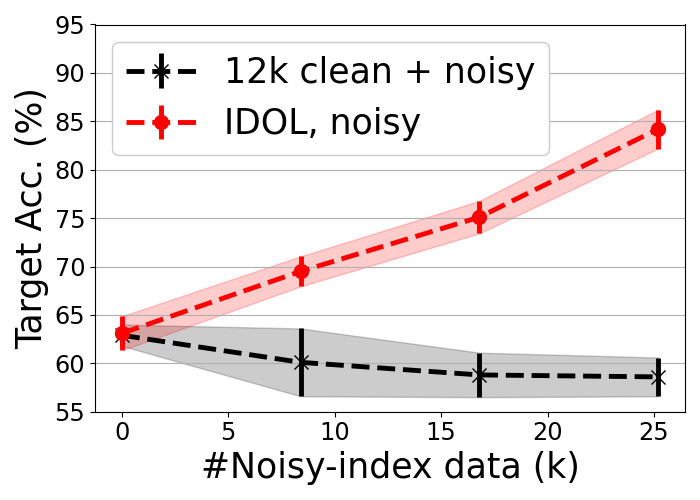}}%
    \endminipage \hfill
    \minipage{0.5\textwidth}
    \mbox{\small \quad\quad\quad(a) Rotated MNIST.}\vfill
    {\includegraphics[width=0.6\textwidth]{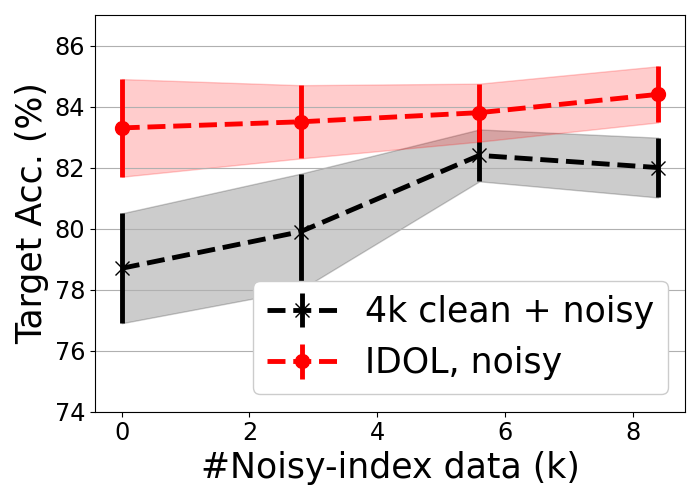}}
    \mbox{\small \quad\quad\quad\quad\quad(b) Portraits.}\vfill
    \endminipage 
    \vspace{-5pt}
    \caption{\small Different numbers of noisily-indexed intermediate data. }  
    \label{fig:unindex_mnist} 
    \vspace{-20pt}
\end{wrapfigure}

\subsection{Case study: learning with low-quality intermediate data or indexes}
In practice, the intermediate unlabeled data or the pre-defined domain indexes for them might not be perfect. We consider several realistic cases to show that our method can make GDA more robust to the quality of the intermediate data or the  domain sequences. First, we investigate \emph{fewer intermediate data}: \autoref{eq:error_bound} implies that the target error increases if the  intermediate data are too sparse. We repeat the experiments in~\autoref{exp:main_study} but with only $30\%$ intermediate data. Second, we consider \emph{noisy indexes}: building upon $30\%$ clean indexes, we further add more data with noisy, random indexes.  

\autoref{fig:unindex_mnist} summarizes the results. We first notice that with only $30\%$ intermediate data, even if the indexes are clean, the accuracy decreases drastically by $14\%$ for MNIST and by $5\%$ for Portraits. For MNIST, adding more noisily-indexed data actually decreases the accuracy. Portraits gains some improvements with the noisily-indexed data, suggesting that the task could benefit from learning with intermediate data (even without accurate indexes). 
One advantage of \IDOL is that we can annotate domain indexes for any unindexed data. On both datasets, \IDOL stably improves with more data available, significantly outperforming that without \IDOL.

\begin{wraptable}{r}{0.35\textwidth}
    \footnotesize
    \vspace{-20pt}
	\caption{\small CIFAR10-STL.} 
	\tabcolsep 1.5pt
	\renewcommand{\arraystretch}{0.75}

	\begin{tabular}{l|c}
	\toprule
	Method & Target Acc.\\
	\midrule
	Source only & 76.6$\pm$0.4 \\
	UDA ($\sT$) (lr $=10^{-4}$) & 69.4$\pm$0.4 \\
	UDA ($\sT$) (lr $=10^{-5}$) & 75.1$\pm$0.3 \\
	UDA ($\sT$ + $\sU$) & 61.1$\pm$0.8 \\
	GDA w/ conf. & 77.1$\pm$0.5 \\
	GDA w/ \IDOL & 78.1$\pm$0.4 \\
    \bottomrule
	\end{tabular}
	\label{tbl:cifar_stl}
	\vspace{-10pt}
\end{wraptable} 

\subsection{Case study: CIFAR10-STL UDA with additional, open-domain unlabeled data}
\label{ss_cifar_exp}
We examine \IDOL on a CIFAR10-to-STL~\cite{krizhevsky2009learning,coates2011analysis} UDA task which is known to be challenging due to the small data size~\cite{cicek2019unsupervised}.
The STL dataset has an additional unlabeled set, whose data are sub-sampled from ImageNet~\cite{deng2009imagenet}. We use this set as the intermediate data, which contains unknown or outlier domains and classes.
Here, we do not aim to compete with the state of the arts on the CIFAR10-to-STL UDA task, but study a more general application scenario for \IDOL. 

To leverage these data, we use $M=3$ and filter out $80\%$ of them using the classifier confidence. We first train a ResNet-20~\cite{he2016deep} source model. In \autoref{tbl:cifar_stl}, we observe that direct UDA using self-training on STL or unlabeled data are inferior to the source model, even if the learning rate is carefully tuned. 
GDA with \IDOL achieves the highest accuracy, demonstrating its robustness and wide applicability.


\section{Related Work}
\label{s_related}

\paragraph{Domain adaptation.}
UDA has been studied extensively~\cite{pan2010domain,ben2010theory,sun2016return}, and many different approaches have been proposed such as minimizing domain divergence~\cite{long2015learning, shu2018dirt, tzeng2014deep,ganin2016domain,gretton2009covariate,sugiyama2007covariate}, cross-domain~\cite{saito2018maximum, saito2017adversarial, lee2019drop,shu2018dirt} and domain-invariant features~\cite{hoffman2017cycada,wang2019transferable,pei2018multi,volpi2018adversarial,xu2020adversarial,yan2020domain,goodfellow2014generative}. Self-training recently emerges as a simple yet effective approach~\cite{chen2011co,kim2019self,zou2018unsupervised,khodabandeh2019robust,tao2018zero,liang2019distant,inoue2018cross}. It uses the source model to provide
pseudo-labels for unlabeled target data, and  
approaches UDA via supervised learning in the target domain~\cite{lee2013pseudo,mcclosky2006effective,mcclosky2006reranking}. Theoretical analyses for applying self-training in UDA are provided in~\cite{wei2020theoretical,chen2020self}.

\paragraph{Gradual domain adaptation.}
Many UDA works show the benefits of gradually bridging the domain gap. With features as input, several work~\cite{gopalan2011domain,gong2012geodesic,cui2014flowing} construct Grassmannian manifolds between the source and the target. For deep models, progressive DA~\cite{hsu2020progressivedet} proposes to create the synthetic intermediate domain with an image-to-image translation network. Other works~\cite{cui2020gradually,gong2019dlow} instead generate intermediate data with jointly-trained generators. \citet{na2020fixbi} augments the intermediate data with Mixup~\cite{zhang2017mixup}. 
Unlike all the above work that focus on building synthetic intermediate domains given only the source and target data, gradual domain adaptation proposed in~\cite{hoffman2014continuous,gadermayr2018gradual,wulfmeier2018incremental,kumar2020understanding} studies how to leverage extra real intermediate data with pre-defined domain indexes. 

\paragraph{Learning with cycle consistency.}
The concept of cycle consistency has been successfully applied in many machine learning tasks. On a high level, cycle supervision enforces that the translation to the target should be able to be translated back to match the source. For instance, back-translation~\cite{sennrich2015improving,he2016dual} is a popular technique to perform unsupervised neural translation in natural language processing. Many computer vision tasks such as image generation~\cite{russo2018source}, image segmentation\cite{li2019bidirectional}, style transfer~\cite{zhu2017unpaired}, etc., can also be improved by matching the structural outputs with cyclic signals. We propose a novel approach that uses cycle consistency to discover intermediate domains for gradual adaptation. 

\paragraph{Discovering feature subspaces.}
We further discuss the connections of our work to existing efforts about subspace learning. One powerful technique is subspace clustering~\cite{kriegel2009clustering,ng2001spectral,elhamifar2013sparse} that partitions data into many subspaces (\eg, classes) unsupervisedly. However, it may not be applied to GDA directly since it groups with discriminative features which might not capture domain shifts. Specifically for UDA, some works aim to discover sub-spaces in the source/target domains~\cite{gong2013reshaping,mancini2018boosting,xiong2014latent,hoffman2012discovering,liu2020open}. The assumption is that the datasets might not be collected from one environment but generated from many different distributions, re-formulating it as a multi-domain adaptation problem. We note that, the sub-domains discovered in these methods are fundamentally different from the domain sequence we discover. The sub-domains are assumed to have a dedicated adaptation mapping to the target domain. The domain sequences in GDA are for constructing a path from the source to the target via intermediate data. How to link these learning techniques to GDA will be interesting future work.


\section{Conclusion}
\label{s_disc} Gradual domain adaptation leverages intermediate data to bridge the domain gap between the source and target domains. However, it relies on a strong premise --- prior knowledge of domain indexes to sort the intermediate data into a domain sequence. We propose the \IDOL algorithm that can produce comparable or even better domain sequences without pre-defined indexes. With a simple progressively-trained domain discriminator, it matches the performance of using pre-defined indexes, and further improves with the proposed cycle-consistency refinement. Essentially, \IDOL is a useful tool to augment any GDA algorithms with high-quality domain sequences given unindexed data.

\section*{Acknowledgments and funding transparency statement}
This research is partially supported by NSF IIS-2107077 and the OSU GI Development funds. We are thankful for the generous support of the computational resources by the Ohio Supercomputer Center.

{\small
\bibliography{main}

\begin{thebibliography}{78}
\providecommand{\natexlab}[1]{#1}
\providecommand{\url}[1]{\texttt{#1}}
\expandafter\ifx\csname urlstyle\endcsname\relax
  \providecommand{\doi}[1]{doi: #1}\else
  \providecommand{\doi}{doi: \begingroup \urlstyle{rm}\Url}\fi

\bibitem[Ben-David et~al.(2010)Ben-David, Blitzer, Crammer, Kulesza, Pereira,
  and Vaughan]{ben2010theory}
Shai Ben-David, John Blitzer, Koby Crammer, Alex Kulesza, Fernando Pereira, and
  Jennifer~Wortman Vaughan.
\newblock A theory of learning from different domains.
\newblock \emph{Machine learning}, 79\penalty0 (1-2):\penalty0 151--175, 2010.

\bibitem[Chen et~al.(2011)Chen, Weinberger, and Blitzer]{chen2011co}
Minmin Chen, Kilian~Q Weinberger, and John Blitzer.
\newblock Co-training for domain adaptation.
\newblock In \emph{NeurIPS}, 2011.

\bibitem[Chen et~al.(2020)Chen, Wei, Kumar, and Ma]{chen2020self}
Yining Chen, Colin Wei, Ananya Kumar, and Tengyu Ma.
\newblock Self-training avoids using spurious features under domain shift.
\newblock In \emph{NeurIPS}, 2020.

\bibitem[Cicek and Soatto(2019)]{cicek2019unsupervised}
Safa Cicek and Stefano Soatto.
\newblock Unsupervised domain adaptation via regularized conditional alignment.
\newblock In \emph{Proceedings of the IEEE/CVF International Conference on
  Computer Vision}, pages 1416--1425, 2019.

\bibitem[Coates et~al.(2011)Coates, Ng, and Lee]{coates2011analysis}
Adam Coates, Andrew Ng, and Honglak Lee.
\newblock An analysis of single-layer networks in unsupervised feature
  learning.
\newblock In \emph{Proceedings of the fourteenth international conference on
  artificial intelligence and statistics}, pages 215--223. JMLR Workshop and
  Conference Proceedings, 2011.

\bibitem[Combes et~al.(2020)Combes, Zhao, Wang, and Gordon]{combes2020domain}
Remi Tachet~des Combes, Han Zhao, Yu-Xiang Wang, and Geoff Gordon.
\newblock Domain adaptation with conditional distribution matching and
  generalized label shift.
\newblock In \emph{NeurIPS}, 2020.

\bibitem[Cui et~al.(2020)Cui, Wang, Zhuo, Su, Huang, and
  Tian]{cui2020gradually}
Shuhao Cui, Shuhui Wang, Junbao Zhuo, Chi Su, Qingming Huang, and Qi~Tian.
\newblock Gradually vanishing bridge for adversarial domain adaptation.
\newblock In \emph{Proceedings of the IEEE/CVF Conference on Computer Vision
  and Pattern Recognition}, pages 12455--12464, 2020.

\bibitem[Cui et~al.(2014)Cui, Li, Xu, Shan, Chen, and Li]{cui2014flowing}
Zhen Cui, Wen Li, Dong Xu, Shiguang Shan, Xilin Chen, and Xuelong Li.
\newblock Flowing on riemannian manifold: Domain adaptation by shifting
  covariance.
\newblock \emph{IEEE transactions on cybernetics}, 44\penalty0 (12):\penalty0
  2264--2273, 2014.

\bibitem[Deng et~al.(2009)Deng, Dong, Socher, Li, Li, and
  Fei-Fei]{deng2009imagenet}
Jia Deng, Wei Dong, Richard Socher, Li-Jia Li, Kai Li, and Li~Fei-Fei.
\newblock Imagenet: A large-scale hierarchical image database.
\newblock In \emph{2009 IEEE conference on computer vision and pattern
  recognition}, pages 248--255. Ieee, 2009.

\bibitem[Elhamifar and Vidal(2013)]{elhamifar2013sparse}
Ehsan Elhamifar and Ren{\'e} Vidal.
\newblock Sparse subspace clustering: Algorithm, theory, and applications.
\newblock \emph{IEEE transactions on pattern analysis and machine
  intelligence}, 35\penalty0 (11):\penalty0 2765--2781, 2013.

\bibitem[Farshchian et~al.(2018)Farshchian, Gallego, Cohen, Bengio, Miller, and
  Solla]{farshchian2018adversarial}
Ali Farshchian, Juan~A Gallego, Joseph~P Cohen, Yoshua Bengio, Lee~E Miller,
  and Sara~A Solla.
\newblock Adversarial domain adaptation for stable brain-machine interfaces.
\newblock \emph{arXiv preprint arXiv:1810.00045}, 2018.

\bibitem[Gadermayr et~al.(2018)Gadermayr, Eschweiler, Klinkhammer, Boor, and
  Merhof]{gadermayr2018gradual}
Michael Gadermayr, Dennis Eschweiler, Barbara~Mara Klinkhammer, Peter Boor, and
  Dorit Merhof.
\newblock Gradual domain adaptation for segmenting whole slide images showing
  pathological variability.
\newblock In \emph{International Conference on Image and Signal Processing},
  pages 461--469. Springer, 2018.

\bibitem[Ganin et~al.(2016)Ganin, Ustinova, Ajakan, Germain, Larochelle,
  Laviolette, Marchand, and Lempitsky]{ganin2016domain}
Yaroslav Ganin, Evgeniya Ustinova, Hana Ajakan, Pascal Germain, Hugo
  Larochelle, Fran{\c{c}}ois Laviolette, Mario Marchand, and Victor Lempitsky.
\newblock Domain-adversarial training of neural networks.
\newblock \emph{JMLR}, 17\penalty0 (1):\penalty0 2096--2030, 2016.

\bibitem[Ginosar et~al.(2015)Ginosar, Rakelly, Sachs, Yin, and
  Efros]{ginosar2015century}
Shiry Ginosar, Kate Rakelly, Sarah Sachs, Brian Yin, and Alexei~A Efros.
\newblock A century of portraits: A visual historical record of american high
  school yearbooks.
\newblock In \emph{Proceedings of the IEEE International Conference on Computer
  Vision Workshops}, pages 1--7, 2015.

\bibitem[Gong et~al.(2012)Gong, Shi, Sha, and Grauman]{gong2012geodesic}
Boqing Gong, Yuan Shi, Fei Sha, and Kristen Grauman.
\newblock Geodesic flow kernel for unsupervised domain adaptation.
\newblock In \emph{2012 IEEE conference on computer vision and pattern
  recognition}, pages 2066--2073. IEEE, 2012.

\bibitem[Gong et~al.(2013)Gong, Grauman, and Sha]{gong2013reshaping}
Boqing Gong, Kristen Grauman, and Fei Sha.
\newblock Reshaping visual datasets for domain adaptation.
\newblock \emph{Advances in Neural Information Processing Systems}, 2013.

\bibitem[Gong et~al.(2019)Gong, Li, Chen, and Gool]{gong2019dlow}
Rui Gong, Wen Li, Yuhua Chen, and Luc~Van Gool.
\newblock Dlow: Domain flow for adaptation and generalization.
\newblock In \emph{Proceedings of the IEEE/CVF Conference on Computer Vision
  and Pattern Recognition}, pages 2477--2486, 2019.

\bibitem[Goodfellow et~al.(2014)Goodfellow, Pouget-Abadie, Mirza, Xu,
  Warde-Farley, Ozair, Courville, and Bengio]{goodfellow2014generative}
Ian Goodfellow, Jean Pouget-Abadie, Mehdi Mirza, Bing Xu, David Warde-Farley,
  Sherjil Ozair, Aaron Courville, and Yoshua Bengio.
\newblock Generative adversarial nets.
\newblock In \emph{Advances in neural information processing systems}, pages
  2672--2680, 2014.

\bibitem[Gopalan et~al.(2011)Gopalan, Li, and Chellappa]{gopalan2011domain}
Raghuraman Gopalan, Ruonan Li, and Rama Chellappa.
\newblock Domain adaptation for object recognition: An unsupervised approach.
\newblock In \emph{2011 international conference on computer vision}, pages
  999--1006. IEEE, 2011.

\bibitem[Gretton et~al.(2009)Gretton, Smola, Huang, Schmittfull, Borgwardt, and
  Sch{\"o}lkopf]{gretton2009covariate}
Arthur Gretton, Alex Smola, Jiayuan Huang, Marcel Schmittfull, Karsten
  Borgwardt, and Bernhard Sch{\"o}lkopf.
\newblock Covariate shift by kernel mean matching.
\newblock \emph{Dataset shift in machine learning}, 3\penalty0 (4):\penalty0 5,
  2009.

\bibitem[Guo et~al.(2017)Guo, Pleiss, Sun, and Weinberger]{guo2017calibration}
Chuan Guo, Geoff Pleiss, Yu~Sun, and Kilian~Q Weinberger.
\newblock On calibration of modern neural networks.
\newblock In \emph{International Conference on Machine Learning}, pages
  1321--1330. PMLR, 2017.

\bibitem[He et~al.(2016{\natexlab{a}})He, Xia, Qin, Wang, Yu, Liu, and
  Ma]{he2016dual}
Di~He, Yingce Xia, Tao Qin, Liwei Wang, Nenghai Yu, Tie-Yan Liu, and Wei-Ying
  Ma.
\newblock Dual learning for machine translation.
\newblock \emph{Advances in neural information processing systems},
  29:\penalty0 820--828, 2016{\natexlab{a}}.

\bibitem[He et~al.(2016{\natexlab{b}})He, Zhang, Ren, and Sun]{he2016deep}
Kaiming He, Xiangyu Zhang, Shaoqing Ren, and Jian Sun.
\newblock Deep residual learning for image recognition.
\newblock In \emph{CVPR}, 2016{\natexlab{b}}.

\bibitem[Hoffman et~al.(2012)Hoffman, Kulis, Darrell, and
  Saenko]{hoffman2012discovering}
Judy Hoffman, Brian Kulis, Trevor Darrell, and Kate Saenko.
\newblock Discovering latent domains for multisource domain adaptation.
\newblock In \emph{European Conference on Computer Vision}, pages 702--715.
  Springer, 2012.

\bibitem[Hoffman et~al.(2014)Hoffman, Darrell, and
  Saenko]{hoffman2014continuous}
Judy Hoffman, Trevor Darrell, and Kate Saenko.
\newblock Continuous manifold based adaptation for evolving visual domains.
\newblock In \emph{Proceedings of the IEEE Conference on Computer Vision and
  Pattern Recognition}, pages 867--874, 2014.

\bibitem[Hoffman et~al.(2018)Hoffman, Tzeng, Park, Zhu, Isola, Saenko, Efros,
  and Darrell]{hoffman2017cycada}
Judy Hoffman, Eric Tzeng, Taesung Park, Jun-Yan Zhu, Phillip Isola, Kate
  Saenko, Alexei~A Efros, and Trevor Darrell.
\newblock Cycada: Cycle-consistent adversarial domain adaptation.
\newblock In \emph{ICML}, 2018.

\bibitem[Hsu et~al.(2020)Hsu, Yao, Tsai, Hung, Tseng, Singh, and
  Yang]{hsu2020progressivedet}
Han-Kai Hsu, Chun-Han Yao, Yi-Hsuan Tsai, Wei-Chih Hung, Hung-Yu Tseng, Maneesh
  Singh, and Ming-Hsuan Yang.
\newblock Progressive domain adaptation for object detection.
\newblock In \emph{IEEE Winter Conference on Applications of Computer Vision
  (WACV)}, 2020.

\bibitem[Inoue et~al.(2018)Inoue, Furuta, Yamasaki, and Aizawa]{inoue2018cross}
Naoto Inoue, Ryosuke Furuta, Toshihiko Yamasaki, and Kiyoharu Aizawa.
\newblock Cross-domain weakly-supervised object detection through progressive
  domain adaptation.
\newblock In \emph{CVPR}, 2018.

\bibitem[Jamal et~al.(2020)Jamal, Brown, Yang, Wang, and
  Gong]{jamal2020rethinking}
Muhammad~Abdullah Jamal, Matthew Brown, Ming-Hsuan Yang, Liqiang Wang, and
  Boqing Gong.
\newblock Rethinking class-balanced methods for long-tailed visual recognition
  from a domain adaptation perspective.
\newblock In \emph{Proceedings of the IEEE/CVF Conference on Computer Vision
  and Pattern Recognition}, pages 7610--7619, 2020.

\bibitem[Khodabandeh et~al.(2019)Khodabandeh, Vahdat, Ranjbar, and
  Macready]{khodabandeh2019robust}
Mehran Khodabandeh, Arash Vahdat, Mani Ranjbar, and William~G Macready.
\newblock A robust learning approach to domain adaptive object detection.
\newblock In \emph{ICCV}, 2019.

\bibitem[Kim et~al.(2019)Kim, Choi, Kim, and Kim]{kim2019self}
Seunghyeon Kim, Jaehoon Choi, Taekyung Kim, and Changick Kim.
\newblock Self-training and adversarial background regularization for
  unsupervised domain adaptive one-stage object detection.
\newblock In \emph{ICCV}, 2019.

\bibitem[Kingma and Ba(2015)]{kingma2014adam}
Diederik~P Kingma and Jimmy Ba.
\newblock Adam: A method for stochastic optimization.
\newblock In \emph{ICLR}, 2015.

\bibitem[Kriegel et~al.(2009)Kriegel, Kr{\"o}ger, and
  Zimek]{kriegel2009clustering}
Hans-Peter Kriegel, Peer Kr{\"o}ger, and Arthur Zimek.
\newblock Clustering high-dimensional data: A survey on subspace clustering,
  pattern-based clustering, and correlation clustering.
\newblock \emph{Acm transactions on knowledge discovery from data (tkdd)},
  3\penalty0 (1):\penalty0 1--58, 2009.

\bibitem[Krizhevsky et~al.(2009)Krizhevsky, Hinton,
  et~al.]{krizhevsky2009learning}
Alex Krizhevsky, Geoffrey Hinton, et~al.
\newblock Learning multiple layers of features from tiny images.
\newblock 2009.

\bibitem[Kumar et~al.(2020)Kumar, Ma, and Liang]{kumar2020understanding}
Ananya Kumar, Tengyu Ma, and Percy Liang.
\newblock Understanding self-training for gradual domain adaptation.
\newblock In \emph{International Conference on Machine Learning}, pages
  5468--5479. PMLR, 2020.

\bibitem[Larochelle et~al.(2007)Larochelle, Erhan, Courville, Bergstra, and
  Bengio]{larochelle2007empirical}
Hugo Larochelle, Dumitru Erhan, Aaron Courville, James Bergstra, and Yoshua
  Bengio.
\newblock An empirical evaluation of deep architectures on problems with many
  factors of variation.
\newblock In \emph{Proceedings of the 24th international conference on Machine
  learning}, pages 473--480, 2007.

\bibitem[LeCun et~al.(1998)LeCun, Bottou, Bengio, and
  Haffner]{lecun1998gradient}
Yann LeCun, L{\'e}on Bottou, Yoshua Bengio, and Patrick Haffner.
\newblock Gradient-based learning applied to document recognition.
\newblock \emph{Proceedings of the IEEE}, 86\penalty0 (11):\penalty0
  2278--2324, 1998.

\bibitem[Lee(2013)]{lee2013pseudo}
Dong-Hyun Lee.
\newblock Pseudo-label: The simple and efficient semi-supervised learning
  method for deep neural networks.
\newblock In \emph{Workshop on challenges in representation learning, ICML},
  2013.

\bibitem[Lee et~al.(2019)Lee, Kim, Kim, and Jeong]{lee2019drop}
Seungmin Lee, Dongwan Kim, Namil Kim, and Seong-Gyun Jeong.
\newblock Drop to adapt: Learning discriminative features for unsupervised
  domain adaptation.
\newblock In \emph{ICCV}, 2019.

\bibitem[Li et~al.(2019)Li, Yuan, and Vasconcelos]{li2019bidirectional}
Yunsheng Li, Lu~Yuan, and Nuno Vasconcelos.
\newblock Bidirectional learning for domain adaptation of semantic
  segmentation.
\newblock In \emph{Proceedings of the IEEE/CVF Conference on Computer Vision
  and Pattern Recognition}, pages 6936--6945, 2019.

\bibitem[Liang et~al.(2019)Liang, He, Sun, and Tan]{liang2019distant}
Jian Liang, Ran He, Zhenan Sun, and Tieniu Tan.
\newblock Distant supervised centroid shift: A simple and efficient approach to
  visual domain adaptation.
\newblock In \emph{CVPR}, 2019.

\bibitem[Liang et~al.(2017)Liang, Li, and Srikant]{liang2017enhancing}
Shiyu Liang, Yixuan Li, and Rayadurgam Srikant.
\newblock Enhancing the reliability of out-of-distribution image detection in
  neural networks.
\newblock \emph{arXiv preprint arXiv:1706.02690}, 2017.

\bibitem[Liu et~al.(2020)Liu, Miao, Pan, Zhan, Lin, Yu, and Gong]{liu2020open}
Ziwei Liu, Zhongqi Miao, Xingang Pan, Xiaohang Zhan, Dahua Lin, Stella~X Yu,
  and Boqing Gong.
\newblock Open compound domain adaptation.
\newblock In \emph{Proceedings of the IEEE/CVF Conference on Computer Vision
  and Pattern Recognition}, pages 12406--12415, 2020.

\bibitem[Long et~al.(2013)Long, Wang, Ding, Sun, and Yu]{long2013transfer}
Mingsheng Long, Jianmin Wang, Guiguang Ding, Jiaguang Sun, and Philip~S Yu.
\newblock Transfer feature learning with joint distribution adaptation.
\newblock In \emph{Proceedings of the IEEE international conference on computer
  vision}, pages 2200--2207, 2013.

\bibitem[Long et~al.(2015)Long, Cao, Wang, and Jordan]{long2015learning}
Mingsheng Long, Yue Cao, Jianmin Wang, and Michael~I Jordan.
\newblock Learning transferable features with deep adaptation networks.
\newblock \emph{arXiv preprint arXiv:1502.02791}, 2015.

\bibitem[Long et~al.(2017)Long, Cao, Wang, and Jordan]{long2017conditional}
Mingsheng Long, Zhangjie Cao, Jianmin Wang, and Michael~I Jordan.
\newblock Conditional adversarial domain adaptation.
\newblock \emph{arXiv preprint arXiv:1705.10667}, 2017.

\bibitem[Mancini et~al.(2018)Mancini, Porzi, Bulo, Caputo, and
  Ricci]{mancini2018boosting}
Massimiliano Mancini, Lorenzo Porzi, Samuel~Rota Bulo, Barbara Caputo, and
  Elisa Ricci.
\newblock Boosting domain adaptation by discovering latent domains.
\newblock In \emph{Proceedings of the IEEE Conference on Computer Vision and
  Pattern Recognition}, pages 3771--3780, 2018.

\bibitem[McClosky et~al.(2006{\natexlab{a}})McClosky, Charniak, and
  Johnson]{mcclosky2006effective}
David McClosky, Eugene Charniak, and Mark Johnson.
\newblock Effective self-training for parsing.
\newblock In \emph{ACL}, 2006{\natexlab{a}}.

\bibitem[McClosky et~al.(2006{\natexlab{b}})McClosky, Charniak, and
  Johnson]{mcclosky2006reranking}
David McClosky, Eugene Charniak, and Mark Johnson.
\newblock Reranking and self-training for parser adaptation.
\newblock In \emph{ACL}, 2006{\natexlab{b}}.

\bibitem[McInnes et~al.(2018)McInnes, Healy, and Melville]{mcinnes2018umap}
Leland McInnes, John Healy, and James Melville.
\newblock Umap: Uniform manifold approximation and projection for dimension
  reduction.
\newblock \emph{arXiv preprint arXiv:1802.03426}, 2018.

\bibitem[Na et~al.(2020)Na, Jung, Chang, and Hwang]{na2020fixbi}
Jaemin Na, Heechul Jung, HyungJin Chang, and Wonjun Hwang.
\newblock Fixbi: Bridging domain spaces for unsupervised domain adaptation.
\newblock \emph{arXiv preprint arXiv:2011.09230}, 2020.

\bibitem[Ng et~al.(2001)Ng, Jordan, and Weiss]{ng2001spectral}
Andrew Ng, Michael Jordan, and Yair Weiss.
\newblock On spectral clustering: Analysis and an algorithm.
\newblock \emph{Advances in neural information processing systems},
  14:\penalty0 849--856, 2001.

\bibitem[Pan et~al.(2010)Pan, Tsang, Kwok, and Yang]{pan2010domain}
Sinno~Jialin Pan, Ivor~W Tsang, James~T Kwok, and Qiang Yang.
\newblock Domain adaptation via transfer component analysis.
\newblock \emph{IEEE Transactions on Neural Networks}, 22\penalty0
  (2):\penalty0 199--210, 2010.

\bibitem[Pei et~al.(2018)Pei, Cao, Long, and Wang]{pei2018multi}
Zhongyi Pei, Zhangjie Cao, Mingsheng Long, and Jianmin Wang.
\newblock Multi-adversarial domain adaptation.
\newblock In \emph{Thirty-Second AAAI Conference on Artificial Intelligence},
  2018.

\bibitem[Ren et~al.(2018)Ren, Zeng, Yang, and Urtasun]{ren2018learning}
Mengye Ren, Wenyuan Zeng, Bin Yang, and Raquel Urtasun.
\newblock Learning to reweight examples for robust deep learning.
\newblock In \emph{International Conference on Machine Learning}, pages
  4334--4343. PMLR, 2018.

\bibitem[Russo et~al.(2018)Russo, Carlucci, Tommasi, and
  Caputo]{russo2018source}
Paolo Russo, Fabio~M Carlucci, Tatiana Tommasi, and Barbara Caputo.
\newblock From source to target and back: symmetric bi-directional adaptive
  gan.
\newblock In \emph{Proceedings of the IEEE Conference on Computer Vision and
  Pattern Recognition}, pages 8099--8108, 2018.

\bibitem[Saito et~al.(2017)Saito, Ushiku, Harada, and
  Saenko]{saito2017adversarial}
Kuniaki Saito, Yoshitaka Ushiku, Tatsuya Harada, and Kate Saenko.
\newblock Adversarial dropout regularization.
\newblock \emph{arXiv preprint arXiv:1711.01575}, 2017.

\bibitem[Saito et~al.(2018)Saito, Watanabe, Ushiku, and
  Harada]{saito2018maximum}
Kuniaki Saito, Kohei Watanabe, Yoshitaka Ushiku, and Tatsuya Harada.
\newblock Maximum classifier discrepancy for unsupervised domain adaptation.
\newblock In \emph{CVPR}, 2018.

\bibitem[Sennrich et~al.(2015)Sennrich, Haddow, and
  Birch]{sennrich2015improving}
Rico Sennrich, Barry Haddow, and Alexandra Birch.
\newblock Improving neural machine translation models with monolingual data.
\newblock \emph{arXiv preprint arXiv:1511.06709}, 2015.

\bibitem[Sethi and Kantardzic(2017)]{sethi2017reliable}
Tegjyot~Singh Sethi and Mehmed Kantardzic.
\newblock On the reliable detection of concept drift from streaming unlabeled
  data.
\newblock \emph{Expert Systems with Applications}, 82:\penalty0 77--99, 2017.

\bibitem[Shu et~al.(2018)Shu, Bui, Narui, and Ermon]{shu2018dirt}
Rui Shu, Hung~H Bui, Hirokazu Narui, and Stefano Ermon.
\newblock A dirt-t approach to unsupervised domain adaptation.
\newblock In \emph{ICLR}, 2018.

\bibitem[Sugiyama et~al.(2007)Sugiyama, Krauledat, and
  M{\~A}{\v{z}}ller]{sugiyama2007covariate}
Masashi Sugiyama, Matthias Krauledat, and Klaus-Robert M{\~A}{\v{z}}ller.
\newblock Covariate shift adaptation by importance weighted cross validation.
\newblock \emph{JMLR}, 8\penalty0 (May):\penalty0 985--1005, 2007.

\bibitem[Sun et~al.(2016)Sun, Feng, and Saenko]{sun2016return}
Baochen Sun, Jiashi Feng, and Kate Saenko.
\newblock Return of frustratingly easy domain adaptation.
\newblock In \emph{AAAI}, 2016.

\bibitem[Tao et~al.(2018)Tao, Yang, and Cai]{tao2018zero}
Qingyi Tao, Hao Yang, and Jianfei Cai.
\newblock Zero-annotation object detection with web knowledge transfer.
\newblock In \emph{ECCV}, 2018.

\bibitem[Tzeng et~al.(2014)Tzeng, Hoffman, Zhang, Saenko, and
  Darrell]{tzeng2014deep}
Eric Tzeng, Judy Hoffman, Ning Zhang, Kate Saenko, and Trevor Darrell.
\newblock Deep domain confusion: Maximizing for domain invariance.
\newblock \emph{arXiv preprint arXiv:1412.3474}, 2014.

\bibitem[Vergara et~al.(2012)Vergara, Vembu, Ayhan, Ryan, Homer, and
  Huerta]{vergara2012chemical}
Alexander Vergara, Shankar Vembu, Tuba Ayhan, Margaret~A Ryan, Margie~L Homer,
  and Ram{\'o}n Huerta.
\newblock Chemical gas sensor drift compensation using classifier ensembles.
\newblock \emph{Sensors and Actuators B: Chemical}, 166:\penalty0 320--329,
  2012.

\bibitem[Volpi et~al.(2018)Volpi, Morerio, Savarese, and
  Murino]{volpi2018adversarial}
Riccardo Volpi, Pietro Morerio, Silvio Savarese, and Vittorio Murino.
\newblock Adversarial feature augmentation for unsupervised domain adaptation.
\newblock In \emph{Proceedings of the IEEE Conference on Computer Vision and
  Pattern Recognition}, pages 5495--5504, 2018.

\bibitem[Wang et~al.(2019)Wang, Li, Ye, Long, and Wang]{wang2019transferable}
Ximei Wang, Liang Li, Weirui Ye, Mingsheng Long, and Jianmin Wang.
\newblock Transferable attention for domain adaptation.
\newblock In \emph{Proceedings of the AAAI Conference on Artificial
  Intelligence}, volume~33, pages 5345--5352, 2019.

\bibitem[Wang et~al.(2020)Wang, Xiangyu, Yurong, Li, Hariharan, Campbell,
  Weinberger, and Wei-Lun]{yan2020domain}
Yan Wang, Chen Xiangyu, You Yurong, Erran~Li Li, Bharath Hariharan, Mark
  Campbell, Kilian~Q. Weinberger, and Chao Wei-Lun.
\newblock Train in germany, test in the usa: Making 3d object detectors
  generalize.
\newblock In \emph{CVPR}, 2020.

\bibitem[Wei et~al.(2021)Wei, Shen, Chen, and Ma]{wei2020theoretical}
Colin Wei, Kendrick Shen, Yining Chen, and Tengyu Ma.
\newblock Theoretical analysis of self-training with deep networks on unlabeled
  data.
\newblock In \emph{ICLR}, 2021.

\bibitem[Wulfmeier et~al.(2018)Wulfmeier, Bewley, and
  Posner]{wulfmeier2018incremental}
Markus Wulfmeier, Alex Bewley, and Ingmar Posner.
\newblock Incremental adversarial domain adaptation for continually changing
  environments.
\newblock In \emph{2018 IEEE International conference on robotics and
  automation (ICRA)}, pages 4489--4495. IEEE, 2018.

\bibitem[Xiong et~al.(2014)Xiong, McCloskey, Hsieh, and Corso]{xiong2014latent}
Caiming Xiong, Scott McCloskey, Shao-Hang Hsieh, and Jason Corso.
\newblock Latent domains modeling for visual domain adaptation.
\newblock In \emph{Proceedings of the AAAI Conference on Artificial
  Intelligence}, volume~28, 2014.

\bibitem[Xu et~al.(2020)Xu, Zhang, Ni, Li, Wang, Tian, and
  Zhang]{xu2020adversarial}
Minghao Xu, Jian Zhang, Bingbing Ni, Teng Li, Chengjie Wang, Qi~Tian, and
  Wenjun Zhang.
\newblock Adversarial domain adaptation with domain mixup.
\newblock In \emph{AAAI}, 2020.

\bibitem[Zhang et~al.(2018)Zhang, Cisse, Dauphin, and
  Lopez-Paz]{zhang2017mixup}
Hongyi Zhang, Moustapha Cisse, Yann~N Dauphin, and David Lopez-Paz.
\newblock mixup: Beyond empirical risk minimization.
\newblock In \emph{ICLR}, 2018.

\bibitem[Zhao et~al.(2019)Zhao, Des~Combes, Zhang, and
  Gordon]{zhao2019learning}
Han Zhao, Remi~Tachet Des~Combes, Kun Zhang, and Geoffrey Gordon.
\newblock On learning invariant representations for domain adaptation.
\newblock In \emph{International Conference on Machine Learning}, pages
  7523--7532. PMLR, 2019.

\bibitem[Zhu et~al.(2017)Zhu, Park, Isola, and Efros]{zhu2017unpaired}
Jun-Yan Zhu, Taesung Park, Phillip Isola, and Alexei~A Efros.
\newblock Unpaired image-to-image translation using cycle-consistent
  adversarial networks.
\newblock In \emph{Proceedings of the IEEE international conference on computer
  vision}, pages 2223--2232, 2017.

\bibitem[Zou et~al.(2018)Zou, Yu, Vijaya~Kumar, and Wang]{zou2018unsupervised}
Yang Zou, Zhiding Yu, BVK Vijaya~Kumar, and Jinsong Wang.
\newblock Unsupervised domain adaptation for semantic segmentation via
  class-balanced self-training.
\newblock In \emph{ECCV}, 2018.

\bibitem[Zou et~al.(2019)Zou, Yu, Liu, Kumar, and Wang]{zou2019confidence}
Yang Zou, Zhiding Yu, Xiaofeng Liu, BVK Kumar, and Jinsong Wang.
\newblock Confidence regularized self-training.
\newblock In \emph{Proceedings of the IEEE/CVF International Conference on
  Computer Vision}, pages 5982--5991, 2019.

\end{thebibliography}
\bibliographystyle{plainnat}
}

\newpage

\section*{\LARGE Supplementary Material}
We provide details omitted in the main paper. 
\begin{itemize}
    \item \autoref{suppl-sec:full}: more details of \IDOL (cf.~\autoref{s_approach} of the main paper).
    \item \autoref{suppl-sec:exp_s}: details of implementations and experimental setups, computation cost, more results for refinement in \IDOL (cf.~\autoref{s_exp} of the main paper).
    \item \autoref{suppl-sec:impact}: broader impact and potential negative societal impacts of our work.
    \item \autoref{suppl-sec:limit}: limitations and future work.
\end{itemize}

\section{Details of \IDOL}
\label{suppl-sec:full}

\subsection{Another view for the coarse-to-fine framework of \IDOL}
Another way to understand the coarse-to-fine framework we proposed is from the optimization perspective. In~\autoref{eq:gda} of the main paper, we provide the overall objective of \IDOL, which is however intractable due to (a) no labeled target data to evaluate the domain sequence and (b) the combinatory nature of searching for a domain sequence. We deal with (a) by~\autoref{eq:reverse_gda} of the main paper. That is, we propose the cycle-consistency loss to measure the quality of the domain sequence, which requires no labeled target data. Since~\autoref{eq:reverse_gda} is still hard to solve, we relax it by a greedy approach. We find one domain at a time in sequence, starting from $\sU_1$ (\ie, the one closest to the source) to $\sU_{M-1}$ (\ie, the one closest to the target). Each sub-problem is described in~\autoref{eq:reverse_gda_one} of the main paper. The relaxation may lead to sub-optimal solutions. Concretely, each sub-problem is not aware of other already selected intermediate domains or other future intermediate domains to be selected. To mitigate this issue, we propose to assign each data point a coarse domain score (\ie,~\autoref{ss_index_func} of the main paper), which serves as the initialization of each sub-problem. 

\subsection{Algorithms}
\begin{figure}[h]
    \centering
    {\includegraphics[width=0.9\textwidth]{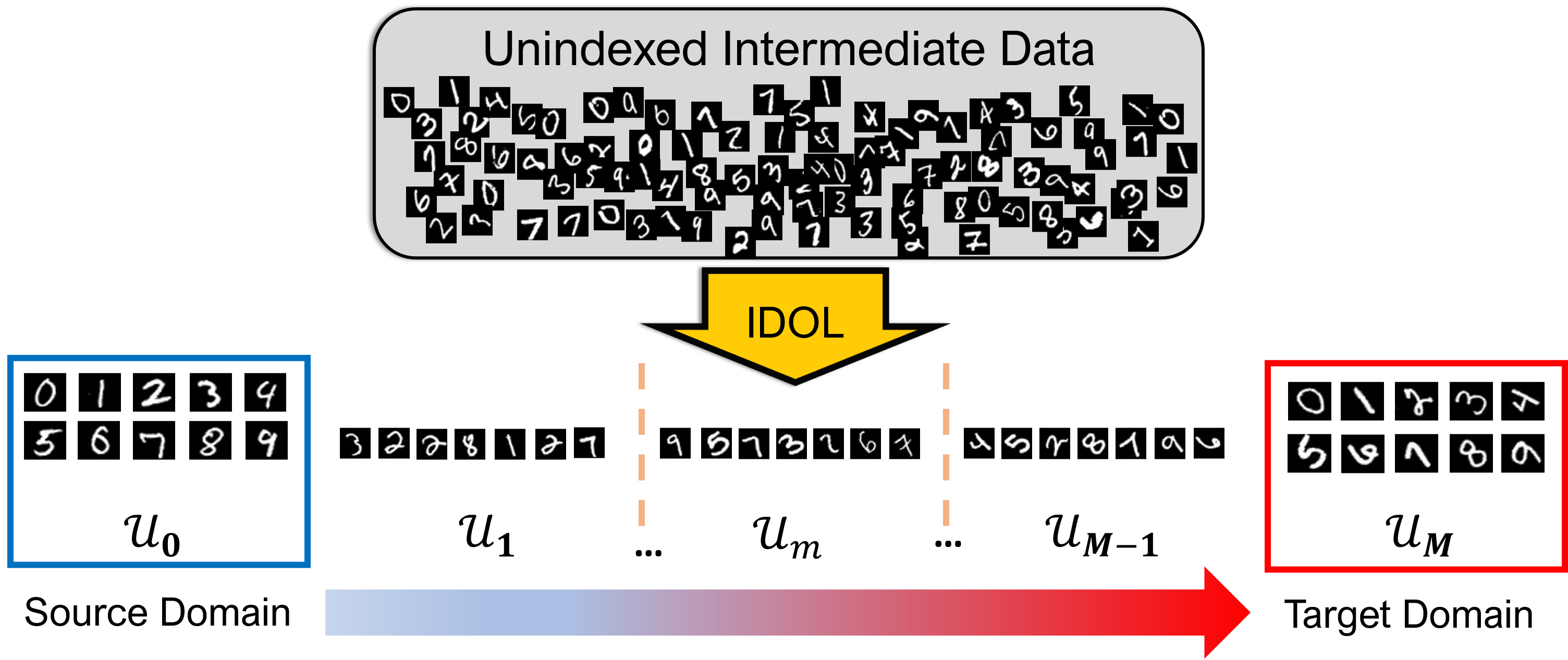}}
    \vspace{-5pt}
    \caption{\small \textbf{Gradual domain adaptation (GDA) without indexed intermediate domains.} Our \IDOL algorithm sorts the unindexed intermediate data into a sequence, from the source domain to the target domain, then it can be further partitioned into several intermediate domains for gradual domain adaptation.}
    \label{fig:split}
\end{figure}

Here we provide the summary of the \IDOL algorithm. \IDOL learns to sort the unindexed intermediate data to a sequence,  partitioned into several intermediate domains. An illustration is provided in~\autoref{fig:split}. As shown in~\autoref{alg:IDOL}, there are three main steps in the overall procedure: first, we construct the coarse domain sequence by learning to predict the domain score for each example and sorting the examples according to the domain scores. Second, we refine the coarse indexes with cycle-consistency as shown in~\autoref{alg:finegrain}. The refinement is decomposed into several steps, gradually discovering the \emph{next} intermediate domain in sequence. Each step is to refine the coarse indexes with meta-reweighting~\cite{ren2018learning,jamal2020rethinking} and takes the closest examples to the current domain as the next domain, as shown in~\autoref{alg:one_step}. Finally, \IDOL outputs a sequence of intermediate data points; it can then be divided into several intermediate domains for gradual domain adaptation.

\begin{algorithm}[H]
\footnotesize
\caption{\textbf{I}ntermediate \textbf{DO}main \textbf{L}abeler (\textbf{\IDOL})}
\label{alg:IDOL}
\KwInput{Labeled source data $\sS$, unlabeled target data $\sT$, intermediate data $\sU$, and \# of domains $M-1$;}
\textbf{Coarse indexing (by progressive training for the domain discriminator):} learn $g(\cdot; \vphi)$ with $\sS, \sT,\sU$ and assign a score $q_i = g(\vx_i^\sU; \vphi)$ to every $\vx_i^\sU\in\sU$ (cf.~\autoref{ss_index_func} in the main paper);
\\
\textbf{Construct:} indexed sequence $I_{\text{coarse}} = (\vx_1, ..., \vx_{|\sU|})$ by sorting $\{q_i = g(\vx_i^\sU; \vphi)|\forall \vx_i^\sU\in\sU\}$; \\
\textbf{Fine-grained indexes:} learn $I_{\text{fine-grained}}$ with refinement (\autoref{alg:finegrain});\\
\textbf{Construct:} domain sequence by chunking $I_{\text{fine-grained}}$ into $M-1$ domains;\\
\KwOut{$(\sU_1, ..., \sU_{M-1})$.}
\end{algorithm}

\begin{algorithm}[h]
    \footnotesize
    \KwInput{Labeled source data $\sS$, \# of domains $M-1$, index sequence $I_{\text{coarse}} = (\vx_1, \ldots, \vx_{|I_{\text{coarse}}|})$.
    }
    \KwInit{Pre-train the source model $\vtheta_0$ on $\sS$, $\sS_0 \leftarrow \sS$, chunk size $C = \frac{|I_{\text{coarse}}|}{M-1}$;}
    \For{$m \in [0, 1, \ldots, M-2]$}
    {
        \KwInit{data parameter $\vq=[\frac{|I_{\text{coarse}}|}{|I_{\text{coarse}}|,}, \frac{|I_{\text{coarse}}|-1}{|I_{\text{coarse}}|}, \ldots, \frac{0}{|I_{\text{coarse}}|}]$;}
        
        Obtain the next domain with  $I_{m+1}\leftarrow$\texttt{FindNextDomain}($\vtheta_m,I_{\text{coarse}},\sS_m, \vq$, $C$);\hfill\text{\color{red}//}\autoref{alg:one_step}\\
        
        Pseudo-label $I_{m+1}$ to construct $\sS_{m+1} = \{(\vx_i, \texttt{sharpen}(f(\vx_i, \vtheta_m)))\}_{\vx_i\in I_{m+1}}$;

        $\vtheta_{m+1}\leftarrow$ self-train $\vtheta_m$ on $\sS_{m+1}$;\\
        
        Update $I_{\text{coarse}} \leftarrow (\vx_i|\vx_i\in I_{\text{coarse}}, \vx_i \notin I_{m+1})$;
    }
    \KwOut{Concatenate $I_{1}, \ldots, I_{M-1}$ as the fine indexes $I_{\text{fine-grained}}$.}
\caption{Refinement of the coarse sequence}
\label{alg:finegrain}
\end{algorithm}

\begin{algorithm}[h]
    \footnotesize
    \KwInput{$\vtheta_m$, (pseudo-)labeled data $\sS_m=\{(\vx_i, y_i)\}_{i=1}^{|\sS_m|}$, intermediate index sequence $I = (\vx_1, ..., \vx_{|I|})$, initial data parameters $\vq\in\R^{|I|}$ ($q_i$ is corresponded to $\vx_i$ in $I$), loss function $\ell$, learning rate $\eta_{\vtheta}, \eta_{\vq}$, and chunk size $C$.
    }
    \While {stop}{
    Detach $\vtheta^{(0)} \leftarrow \vtheta_m$;\\
    \For{$t \in [1, \ldots, T]$}
    {
    Sample a mini-batch $B_{\sU}$ from $I,\vq$;\hfill\text{\color{red}//Forward adaptation.}\\
    $\vtheta^{(t)} \leftarrow \vtheta^{(t-1)} - \eta_{\vtheta} \nabla_{\vtheta^{(t-1)}} \sum_{i\in B_{\sU}} q_i\times\ell(f(\vx_i; \vtheta^{(t-1)}), \texttt{sharpen}(f(\vx_i; \vtheta_m)) )$;\\
    }
    Detach $\vtheta'\leftarrow\vtheta^{(T)}(\vq)$;\\
    \For{$t \in [T, \ldots, 2T-1]$}
    {
    Sample a mini-batch $B_{\sS_m}$ from $\sS$;\hfill\text{\color{red}//Backward adaptation.}\\
    $\vtheta^{(t+1)} \leftarrow \vtheta^{(t)} - \eta_{\vtheta} \nabla_{\vtheta^{(t)}} \frac{1}{|B_{\sS_m}|}\sum_{i\in B_{\sS_m}} \ell(f(\vx_i; \vtheta^{(t)}), \texttt{sharpen}(f(\vx_i; \vtheta')) )$;\\
    }
    Sample a mini-batch $B_{\sS_m}$ from $\sS_m;$\hfill\text{\color{red}//Update data parameters with cycle-consistency.}\\
    Update $\vq\leftarrow \vq-\eta_{\vq} \nabla_{\vq} \frac{1}{|B_{\sS_m}|}\sum_{i\in B_{\sS_m}}\ell(f(\vx_i; \vtheta^{(2T)}(\vq)), y_i)$;\\
    $q_i\leftarrow \max\{0, q_i\}, \forall q_i \in \vq$;\\
    }
    $I\leftarrow$ sort $I$ by $\vq$ (descending order);\\
    \KwOut{the next domain $I[:C]$.}
\caption{Finding the next domain with cycle-consistency (\texttt{FindNextDomain})}
\label{alg:one_step}
\end{algorithm}

\section{Implementation Details}
\label{suppl-sec:exp_s}

\begin{table}[t] 
    \footnotesize
    \small
	\centering
	\tabcolsep 4.5pt
	\renewcommand{\arraystretch}{0.7}
	\caption{\small \IDOL with refinement on different coarse domain scores.}
	\begin{tabular}{l|cc|c|cc}
	\toprule
	Coarse scores & Indexed? & Adaptation & Refined?  & Rotated MNIST & Portraits\\
    \midrule
    \multirow{2}{*}{Classifier confidence} & \multirow{8}{*}{\xmark} & \multirow{8}{*}{GDA} & \xmark & 45.5$\pm$3.5 & 79.3$\pm$1.7\\
    & & & \cmark & 62.5$\pm$2.1 & 83.6$\pm$1.6\\
    \multirow{2}{*}{Manifold distance} &  & &\xmark & 72.4$\pm$3.1 & 81.9$\pm$0.8\\
     &  & &\cmark & 82.4$\pm$2.3 & 85.2$\pm$0.9\\
    \multirow{2}{*}{Domain discriminator} &  & &\xmark& 82.1$\pm$2.7 & 82.3$\pm$0.9\\
     &  & &\cmark& 86.2$\pm$2.2 & 85.1$\pm$1.3\\
    \multirow{2}{*}{Progressive domain discriminator} &   & & \xmark& 85.7$\pm$2.7 & 82.3$\pm$0.9\\
    &   & & \cmark& 87.5$\pm$2.0 & 85.5$\pm$1.0\\
    \bottomrule
	\end{tabular}
	\vspace{0.1cm} 
	
	\label{tbl:main_refine}
	\vspace{-0.6cm} 
\end{table}

\subsection{Experimental setup}
\paragraph{Dataset, model, and optimizer.} The Rotated MNIST and Portraits datasets are resized to $28\times28$ and $32\times32$, respectively, without data augmentation. In the CIAFR10-STL experiments, images are resized to $32\times32$. We use the standard normalization and data augmentation with random horizontal flipping and random cropping as in~\cite{he2016deep}. 

For the Rotated MNIST and Portraits datasets, we adopt the same network used in~\cite{kumar2020understanding}. The network consists of $3$ convolutional layers, each with the kernel size of $5$, stride $2$, and $32$ channel size. We use ReLU activations for all hidden layers. After the convolutional layers, it follows with a dropout layer with $0.5$ dropping rate, a batch-norm layer, and the Softmax classifier. We use the Adam optimizer~\cite{kingma2014adam} with the learning rate $0.001$, batch size $32$, and weight decay $0.02$. We use this optimizer and train for $20$ epochs for the Rotated MNIST and Portraits datasets as the default if not specified, including training the source model, self-training adaptation on each domain, our domain discriminator, and progressive training for each step. 

For the CIFAR10-STL experiments, we train a ResNet-20 for $200$ epochs for the source model and $80$ epochs for both the domain discriminator and progressive training for each step with Adam optimizer with learning rate $0.00001$, batch size $128$, and weight decay $0.0001$. 

We use the same network for our domain discriminator but replace the classifier with a Sigmoid binary classifier. 

\paragraph{GDA.} For GDA, we focus on gradual self-training studied in~\cite{kumar2020understanding}. We follow the common practice to filter out low confidence pseudo-labeled data for self-training on every domain. That is, in applying~\autoref{eq:self-training} of the main paper for self-training, we only use data with high prediction confidences.
We keep top $90\%$ confident examples for the Rotated MNIST and Portraits datasets and top $20\%$ confident examples for the CIFAR10-STL experiments due to the fact that most of the unlabeled data are noisy and could be out-of-distribution. Each domain is trained for $20$ epochs for the Rotated MNIST and Portraits datasets. We train $80$ epochs for the CIFAR10-STL experiments. 

\paragraph{\IDOL.} For hyperparameters specific to \IDOL, we tune with the target validation set. We did not change hyperparameters in GDA for fair comparison. We set $K=2M$ rounds for progressive training. For~\autoref{alg:one_step}, we set $T=10$ and train for in total $30$ epochs for all datasets, with batch size $128$. The learning rates are set as $\eta_{\vtheta} = \eta_{\vq} = 0.001$.

\subsection{Computation cost}
We run our experiments on one GeForce RTX 2080 Ti GPU with Intel i9-9960X CPUs. In~\autoref{alg:one_step}, updating the data parameter $\vq$ requires it to backward on gradients of $\vtheta$, which approximately takes three-time of the computation time of a standard forward-backward pass~\cite{ren2018learning}. 

We estimated the time consumption on the Portraits dataset experiments. For coarse scores, the confidence is simply by applying the classifier to the input data. Calculating the manifold distance via~\cite{mcinnes2018umap} takes about 20 seconds with Intel i9-9960X CPU. The domain discriminator (both w/ or w/o progressive training) and the fine stage involve GPU computation. Using a GeForce RTX 2080 Ti GPU, training a domain discriminator takes about 4 seconds. Progressive training takes about $M$ times longer, where $M$ is the number of intermediate domains (which is $7$ here). This is because it trains the domain discriminator for $M$ rounds. For the refinement stage proposed in cf.~\autoref{ss_metaGDA} based on meta-learning, Discovering one intermediate domain takes about $44$ seconds and we perform it for $M$ rounds.

\subsection{More results on refinement}
We provide the full experiments (cf.~\autoref{tbl:main} in the main paper) of \IDOL refinement on different coarse domain scores in~\autoref{tbl:main_refine}.  We observe the refinement consistently helps on the coarse domain scores, and the quality of the coarse domain scores is important to the final performance. 

Next, we design the following experiment to investigate the variations of different trials and the effects of the number of intermediate samples. We apply \IDOL (with progressive discriminator and refinement) for $10$ runs with different random seeds, and record the intermediate domain index (from source $= 0$ to the target $= M$) to which each sample is assigned. For instance, if a sample is assigned to the second intermediate domain, we record $2$ for this sample. After $10$ runs, each sample will end up with $10$ indices, and we calculate the variance. If a sample is always assigned to the same intermediate domain, the variance will be zero for this sample. We repeat the experiments with $100\%/50\%/25\%$ of intermediate data (dropped randomly). The averaged variance over all samples for both the Rotated MNIST dataset ($M = 19$) and Portraits dataset ($M = 7$). As shown in~\autoref{sup-tbl:var}, the domain sequences formed in different trials are pretty consistent since the variances are small. Rotated MNIST has a higher variance than Portraits probably because the number of intermediate domains is larger. Besides, the amount of the intermediate data does not significantly affect the consistency among trials.

\begin{table}
    \centering
	\caption{\small Variances of domain assignments.} 
	\tabcolsep 1.5pt
	\renewcommand{\arraystretch}{0.75}
	\begin{tabular}{l|cc}
	\toprule
	Intermediate data &	Rotated MNIST &	Portraits\\
	\midrule
	100\% &	1.12&	0.147 \\
    50\%&	1.33&	0.150 \\
    25\%&	1.35&	0.136 \\
    \bottomrule
	\end{tabular}
	\label{sup-tbl:var}
	\vspace{-10pt}
\end{table}

\section{Broader Impact}
\label{suppl-sec:impact}
Unsupervised domain adaptation aims to address the distribution shift from the labeled training data to the unlabeled target data. Our work focuses on leveraging extra unlabeled data to help unsupervised domain adaptation. We only assume that the unlabeled data are generated from underlying distributions that can be viewed as structured shifts from the source to the target domain. To the best of our knowledge, our algorithm would not cause negative effects on data privacy, fairness, security, or other societal impacts.

\section{Limitations and Future Work}
\label{suppl-sec:limit}
Nearly all the domain adaptation algorithms need to make assumptions on the underlying data distributions. For our algorithm \IDOL, we generally require the additional unlabeled data to distribute in between the source and the target domains, which is the standard setup of gradual domain adaptation (cf.~\autoref{s_background} of the main paper). GDA will be less effective if the additional unlabeled data besides the source and target domains do not gradually bridge the two domains or contain outliers. If there are unlabeled data close to the source or target domains but are not distributed in between of them (e.g., if the source/target are digits with $60$/$0$-degree rotation, then digits with $70$-degree are one of such cases), \IDOL may still include those data into the intermediate domains and could potentially degrade the overall performance. In~\autoref{ss_cifar_exp} of the main paper, we further show that, even if the additional unlabeled data contain outliers or does not smoothly bridge the two domains, \IDOL can still effectively leverage the data to improve the performance on the target domain.

Our current scope of the experiments is mainly limited by the availability of benchmark datasets for gradual domain adaptation. This is mainly because GDA is a relatively new setting. Most of the existing datasets for domain adaptation are designed for the conventional unsupervised domain adaptation, in which only the source and target domain data are provided. Given (1) the superior performance of gradual domain adaptation by leveraging the additional unlabeled data and (2) the fact that real-world data change gradually more often than abruptly, we believe it will be useful to develop more datasets for gradual domain adaptation. For instance, datasets in autonomous driving that involve shifts in time and geo-locations could be an option. Other examples include sensor measurements drift over time, evolving road conditions in self-driving cars, and neural signals received by brain-machine interfaces, as pointed out in the introduction of \cite{kumar2020understanding}.

\end{document}